\documentclass[10pt,journal,compsoc]{IEEEtran}
\renewcommand{\baselinestretch}{0.98}
\ifCLASSOPTIONcompsoc
  \usepackage[nocompress]{cite}
\else
  \usepackage{cite}
\fi

\usepackage{graphicx}
\usepackage{comment}
\usepackage{amsmath,amssymb} 
\usepackage{color}

\usepackage{multirow}
\usepackage{multicol}
\usepackage{floatrow}
\usepackage[noend]{algpseudocode}
\usepackage{algorithmicx,algorithm}
\usepackage{bbding}

\usepackage{tikz,xcolor,hyperref}
\usepackage{subfigure}
\usepackage{booktabs}
\usepackage{makecell}
\usepackage{xspace}

\newcommand{\h}{0mm}
\newcommand{\hh}{0mm}

\newcommand{\tabincell}[2]{\begin{tabular}{@{}#1@{}}#2\end{tabular}}
\definecolor{lime}{HTML}{A6CE39}
\DeclareRobustCommand{\orcidicon}{%
    \begin{tikzpicture}
    \draw[lime, fill=lime] (0,0) 
    circle [radius=0.16] 
    node[white] {{\fontfamily{qag}\selectfont \tiny ID}};    \draw[white, fill=white] (-0.0625,0.095) 
    circle [radius=0.007];    \end{tikzpicture}
    \hspace{-2mm}}
\foreach \x in {A, ..., Z}{%
    \expandafter\xdef\csname orcid\x\endcsname{\noexpand\href{https://orcid.org/\csname orcidauthor\x\endcsname}{\noexpand\orcidicon}}
    }

\def \alambic {\includegraphics[width=0.02\linewidth]{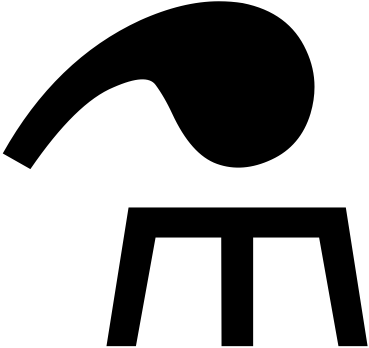}\xspace}

\begin{document}
%
\title{TransCL: Transformer Makes Strong and Flexible Compressive Learning}
%
%
%
%

\author{Chong~Mou, Jian~Zhang
\IEEEcompsocitemizethanks{\IEEEcompsocthanksitem C.~Mou and J.~Zhang are with the School of Electronic and Computer Engineering, Peking University Shenzhen Graduate School, Shenzhen 518055, China. Jian Zhang is also with the Peng Cheng Laboratory, Shenzhen, China. (Corresponding author: Jian Zhang.)\protect\\
E-mail: eechongm@stu.pku.edu.cn; zhangjian.sz@pku.edu.cn
}
\thanks{Manuscript received August 26, 2021; revised December 10, 2021, March 24, 2022 and July 6, 2022; accepted July 14, 2022. This work was supported by National Natural Science Foundation of China (61902009).}}

%
%

\markboth{Journal of \LaTeX\ 2021}%
{Shell \MakeLowercase{\textit{et al.}}: Bare Demo of IEEEtran.cls for Computer Society Journals}
%



\IEEEtitleabstractindextext{%
\begin{abstract}
Compressive learning (CL) is an emerging framework that integrates signal acquisition via compressed sensing (CS) and machine learning for inference tasks directly on a small number of measurements. It can be a promising alternative to classical image-domain methods and enjoys great advantages in memory saving and computational efficiency. However, previous attempts on CL are not only limited to a fixed CS ratio, which lacks flexibility, but also limited to MNIST/CIFAR-like datasets and do not scale to complex real-world high-resolution (HR) data or vision tasks. In this paper, a novel transformer-based compressive learning framework on large-scale images with arbitrary CS ratios, dubbed TransCL, is proposed. Specifically, TransCL first utilizes the strategy of learnable block-based compressed sensing and proposes a flexible linear projection strategy to enable CL to be performed on large-scale images in an efficient block-by-block manner with arbitrary CS ratios. Then, regarding CS measurements from all blocks as a sequence, a pure transformer-based backbone is deployed to perform vision tasks with various task-oriented heads. Our sufficient analysis presents that TransCL exhibits strong resistance to interference and robust adaptability to arbitrary CS ratios. Extensive experiments for complex HR data demonstrate that the proposed TransCL can achieve state-of-the-art performance in image classification and semantic segmentation tasks. In particular, TransCL with a CS ratio of $10\%$ can obtain almost the same performance as when operating directly on the original data and can still obtain satisfying performance even with an extremely low CS ratio of $1\%$. The source codes of our proposed TransCL is available at \url{https://github.com/MC-E/TransCL/}. 
\end{abstract}

}

\maketitle

\IEEEdisplaynontitleabstractindextext

%
\IEEEpeerreviewmaketitle

\IEEEraisesectionheading{\section{Introduction}\label{sec:introduction}}
\IEEEPARstart{S}{ignal} acquisition and processing methods have changed substantially over the last several decades. Many methods have moved from the analog to the digital domain, creating sensing systems that are more robust, flexible, and cost-effective than their analog counterparts. However, in many important real-world applications, the resulting Nyquist rate is so high that it is not viable or even physically impossible. Compressed sensing (CS), built upon the groundbreaking works by Candes \textit{et al.}~\cite{cs0} and Donoho \textit{et al.}~\cite{cs}, has become a more efficient and hardware-friendly signal acquisition method and has been receiving more and more attention. CS theory shows that when a signal has a sparse representation on a certain basis, one can vastly reduce the number of required samples—below the Nyquist rate and can still recover the signal perfectly from its CS measurement. CS has been applied in many practical applications, including but not limited to single-pixel imaging~\cite{spc,spc2}, accelerating magnetic resonance imaging (MRI)~\cite{mri}, wireless telemonitoring~\cite{zhang2012compressed}, cognitive radio communication~\cite{sharma2016application}, and snap compressive imaging \cite{wang2021metasci, wu2021SCI3D}. Correspondingly, there exist several types of CS equipments, \textit{e.g.}, single-pixel camera~\cite{spc}, flexible voxels camera~\cite{fvc}, and  P2C2 camera~\cite{p2c2}, which can project images from the high-dimensional image domain to the low-dimensional measurement domain.

Most of the recent works \cite{cs_w1,cs_w2,cs_w3,cs_w4, zhao2014image,zhao2016video,you2021coast,song2021memory} in the CS community focus on how to reconstruct the original signal, and less attention has been devoted to whether one can perform high-level vision tasks direct in the measurement domain. However, in many applications, such as classification and segmentation, we are not interested in obtaining a precise reconstruction of the scene under view, but rather are only interested in the results of inference tasks. There also exist coupling problems between the reconstruction results and the inference model trained on natural images. Moreover, with the development of deep learning, data privacy has gradually become a concern \cite{priv,priv2}. Thus, in some specific applications, signal reconstruction is undesirable since this step can potentially disclose private information. Directly performing inference tasks in the measurement domain without knowing the sampling matrix makes it almost inaccessible to original visual perception, which is an appropriate scheme to protect data privacy during network training and inference. 

Calderbank~\textit{et al.}~\cite{cl2} and Davenport~\textit{et al.}~\cite{cl1} first construct the concept of compressive learning (CL) in which the inference system is directly built on the top of CS measurements without an explicit reconstruction step. Some theoretical demonstrations of CL have been made in \cite{td1,td2}. Some early works have been made in \cite{cl_w1,cl_w2,cl_w3,cl_w4}. The very recent researches (Tran~\textit{et al.}~\cite{cl_w5,mcl_v2}) propose an efficient method to exploit the multidimensional property of CL. Although there have been several novel attempts, some intractable problems make existing compressive learning methods not yet comparable to the image-domain methods. \textbf{First}, the information loss during the CS sampling process due to the lax sparsity of natural images makes the performance of CL significantly lower than image-domain methods. \textbf{Second}, a large-scale sampling matrix and large matrix multiplication are needed to project real-world images, making the sampling process complicated. The high complexity in CS of high-resolution images is the reason of limiting existing CL methods \cite{cl_w1,cl_w2,cl_w3,cl_w4,cl_w5,mcl_v2} to MNIST/CIFAR-like datasets \cite{mnist,cifar}. \textbf{Third}, different CS ratios will generate measurement vectors with different lengths, leading to the fact that most CL methods can only perform inference tasks with a fixed CS ratio, limiting their practice in some real-world applications. In this paper, we propose several solutions to solve these weaknesses, helping bridge the gap between CL and image-domain methods. 

In fact, an image and its CS measurement have a lot in common, and several helpful priors in the image domain can also be applied in the measurement domain. For instance, the non-local self-similarity prior, which assumes that similar content would recur across the whole scene. This prior has been widely used in the community of image reconstruction~\cite{nonm,BM3D}. However, how to combine this effective prior naturally into CL to resist the information loss has been hardly studied. Recently, Transformer (Vaswani \textit{et al.}~\cite{attis}) has attracted more and more attention from the community of computer vision. This family of networks~\cite{attis,bert,gpt,attis2}, \textit{i.e.}, transformer networks, originate from machine translation and require  serialized inputs, and are very good at modeling long-range correlations in sequences. Thus, transformer can be applied in CL to compensate for information loss through element-wise correlations. In order to make complex real-world data tractable, we propose to adopt the strategy of block-based compressed sensing (BCS) \cite{bcs,bcs2} into CL, which applies CS to image blocks through the same operator to generate a sequence of CS measurements instead of an entire one. BCS is mostly suitable for processing very large-scale images in CL, while requiring significantly less memory to store the sensing matrix and provides a serialized input to the transformer layer. Thus, a significant finding is that BCS and Transformer can be well combined to solve the two main challenges faced by existing CL methods, which greatly inspires our proposed TransCL framework. We further propose a flexible linear projection to enable a single model to handle input measurements with arbitrary CS ratios. In addition, our TransCL exploits a learnable binary sampling matrix, which is more practical and hardware-friendly. 

\textbf{The main contributions of this paper are summarized as follows:}
\begin{itemize}
\item We reformulate the CL problem from a sequence-to-sequence sensing and learning perspective and propose a novel transformer-based CL framework, dubbed TransCL. Our TransCL can be generalized to more complex high-level vision tasks and real-world benchmarks, which helps bridge the gap between inference tasks in the image and measurement domains.
\item We propose a flexible projection scheme to enable a single model to sample and infer images with arbitrary CS ratios. In this way, our method can be trained only once to deal with general cases.
\item We propose incorporating the strategy of learnable block-based compressed sensing (BCS) and binary sampling matrix into CL, making the whole system more memory-saving and hardware-friendly.
\item Extensive experiments demonstrate that our TransCL can achieve state-of-the-art performance in various compressive learning tasks (\textit{i.e.}, image classification and semantic segmentation), especially at extremely low CS ratios (\textit{e.g.}, 1\%). We further verify the robustness of our TransCL in the face of disturbances. 
\end{itemize}

\begin{figure}
    \centering
    \includegraphics[width=1\columnwidth]{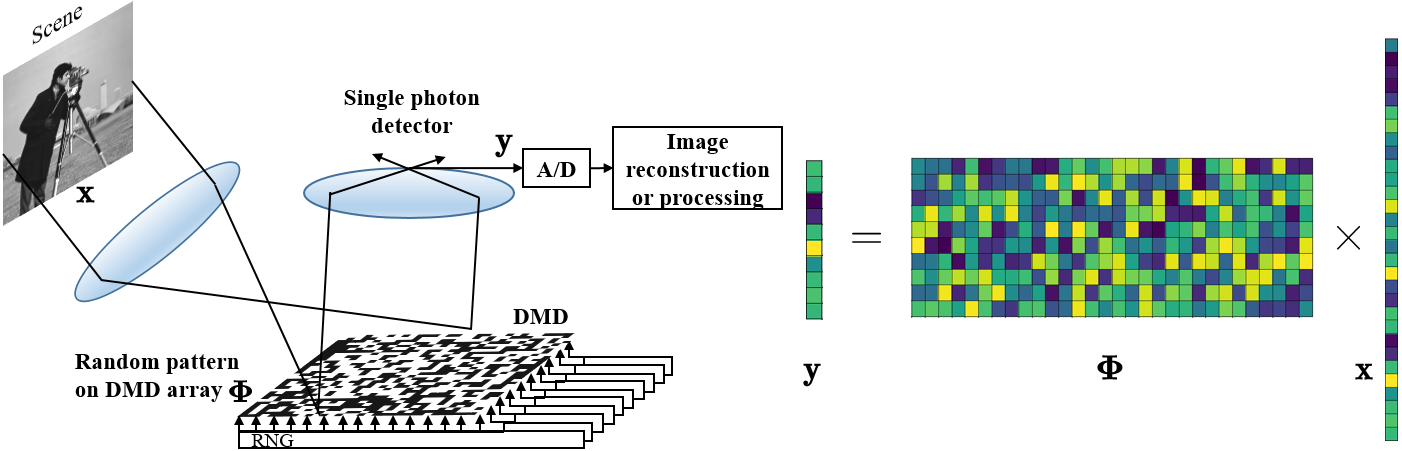}
    \caption{Illustration of CS acquisition for a still image with a single-pixel camera \cite{cs}, which can be formulated as $\mathbf{y}=\mathbf{\Phi}\mathbf{x}$.}
    \label{sampling}
\end{figure}

\section{Related Work}
Our proposed TransCL is closely related to existing compressed sensing methods, compressive learning methods, and transformer networks. Thus, in what follows, we give a brief review of these aspects and focus on the specific methods most relevant to our own.

\subsection{Compressed Sensing}
\label{section_cs}
Compressed sensing (CS) \cite{cs} is an emerging mathematical paradigm that permits, under certain conditions, linear projection of a signal into a dimension much lower than that of the original signal while allowing exact recovery of the signal from its projection. That is, from much fewer acquired measurements than determined by Nyquist sampling theory, CS theory demonstrates that a signal can be reconstructed with high probability when it exhibits sparsity in some transform domain. This novel acquisition strategy is much more hardware-friendly and enables image or video acquisition with a sub-Nyquist sampling rate. By leveraging the redundancy inherent to a signal, CS conducts sampling and compression at the same time, which greatly alleviates the need for high transmission bandwidth and large storage space, enabling low-cost on-sensor data compression.

Concretely, suppose we have the CS measurement  $\mathbf{y}$$\in$$\mathbb{R}^{M}$ of the signal $\mathbf{x}\in \mathbb{R}^{N}$, \textit{i.e.}, $\mathbf{y}=\mathbf{\Phi x}$, where $M \ll N$, $\mathbf{\Phi}\in \mathbb{R}^{M\times N}$ is the sampling matrix, and CS ratio is defined as $\gamma=\frac{M}{N}$. Because the number of unknowns is much larger than the number of observations, reconstructing $\mathbf{x}$ from its corresponding $\mathbf{y}$ is impossible in general; however, if $\mathbf{x}$ is known to be sufficiently sparse in some domain, the exact recovery of $\mathbf{x}$ is possible——that is the fundamental tenet of CS theory. Fig.~\ref{sampling} illustrates the CS acquisition process of a still image with a single-pixel camera. 

Since the advent of the CS, most methods \cite{cs_w1,cs_w2,cs_w3,cs_w4} focus on the inverse problem of CS, which aims to fully recover the signal from its CS measurement. Given the linear measurement $\mathbf{y}$, traditional reconstruction methods usually restore the original signal by solving the following optimization problem:
\begin{equation}
    \mathop{min}\limits_{\mathbf{{x}}}\frac{1}{2}||\mathbf{\Phi}\mathbf{{x}}-\mathbf{y}||^2_2+\lambda||\psi(\mathbf{x})||_1,
\end{equation}
where $\psi (\mathbf{x})$ represents the transformation of $\mathbf{x}$ with respect to some transform operator $\psi(\cdot)$, and the sparsity of $\psi(\mathbf{x})$ is encouraged by the $\ell_1$ norm with $\lambda$ being the regularization parameter (generally pre-defined). However, in many high-level applications, we are not interested in obtaining a precise reconstruction of the scene under view, but rather are the results of specific inference tasks.

\subsection{Compressive Learning}
\begin{figure}[t]
    \centering
    \includegraphics[width=1\columnwidth,height=2.7cm]{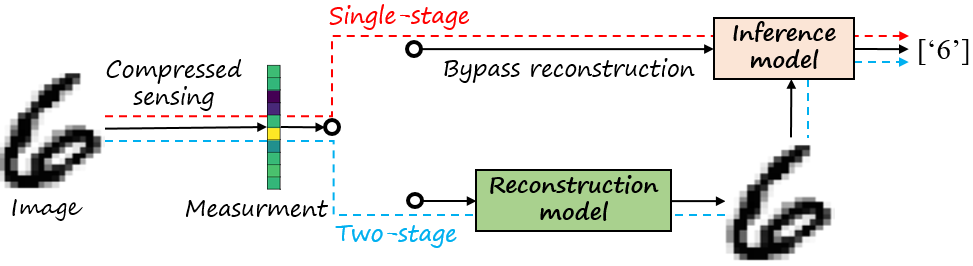}
    \caption{Illustration of two ways of performing inference tasks (i.e., image classification) on CS measurements. The blue dashed line presents the conventional two-stage operation (reconstruction first and then inference in the image domain), and the red dashed line represents the single-stage compressive learning (CL) direct in the measurement domain.}
    \label{cl}
\end{figure}

As illustrated by the red dashed line in Fig.~\ref{cl}, compressive learning (CL) aims to directly perform inference tasks in the measurement domain without signal reconstruction. The concept of CL is first constructed in \cite{cl1,cl2}. This strategy bypasses the reconstruction process and greatly reduces the amount of data. Thus, it is highly cost-effective in terms of storage memory and computational complexity. Moreover, it can help to protect data privacy. The theoretical basics of CL have been well established. The early research \cite{cl2} demonstrates that given certain conditions of the sampling matrix $\mathbf{\Phi}$, the performance of a linear support vector machine (SVM) trained on CS measurements is as good as the best linear threshold classifier trained on the original signal. \cite{td1} proves that the Kullback-Leibler and Chernoff distances between two probability density functions are preserved up to a factor of $\frac{M}{N}$, where $M$ and $N$ represent the length of the CS measurement and original signal, respectively. \cite{td2} exploits the performance bound under perturbation bringing from the uncertainty of the measurement matrix. Some early works \cite{cl_w1,cl_w2,cl_w3,cl_w4} study directly performing inference tasks in the measurement domain. \cite{pamicl1} applies CL to action recognition. With the advances in computing hardware and stochastic optimization techniques, \cite{op1,op2} propose to jointly optimize the sampling matrix with the whole network in an end-to-end manner. Benefiting from the programmable CS methods~\cite{pcs1,pcs2}, the learned sampling matrix has the opportunity to be deployed to hardware for CS. The very recent researches \cite{cl_w5,mcl_v2} propose an efficient method to exploit the multidimensional property of CL. However, existing CL methods are restricted to a small range of high-level vision tasks (\textit{e.g.}, image classification/recognition). These methods are usually difficult or unstable~\cite{cl_ind} to handle large-scale images, and most works focus on small datasets, such as MNIST~\cite{mnist} and CIFAR~\cite{cifar}, not comparable to complex real-world high-resolution data. Furthermore, these CL methods only handle inference tasks with a fixed CS ratio without considering the variation of CS ratio in practical applications.

\begin{figure}[t]
    \centering
    \includegraphics[width=.8\columnwidth,height=3.8cm]{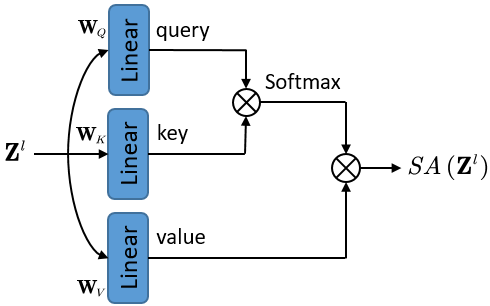}
    \caption{Illustration of the self-attention (SA) mechanism \cite{attis}, which is used to construct long-range correlations among the input sequence.}
    \label{nl}
\end{figure}

\begin{figure*}[h]
    \centering
    \includegraphics[width=.95\linewidth]{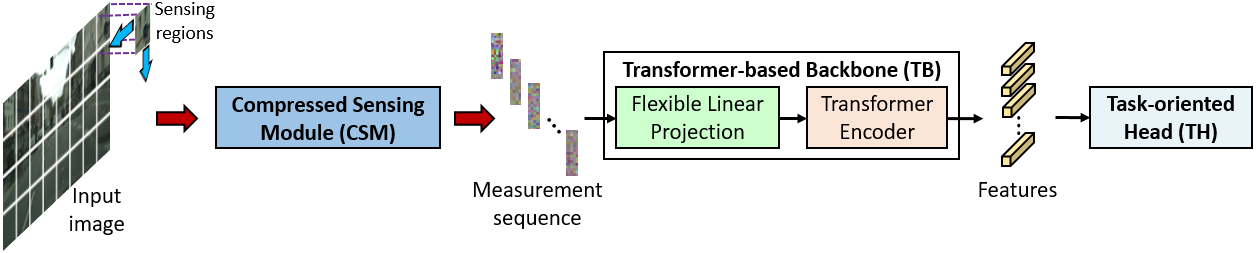}
    \caption{An overview of our proposed transformer-based compressive learning (TransCL) framework, which is composed of compressed sensing module (CSM), Transformer-based Backbone (TB) and Task-oriented Head (TH).}
    \label{net_all}
\end{figure*}

\begin{figure*}[h]
\centering
\small 
\begin{minipage}{0.58\linewidth}
\centering
\includegraphics[width=1\columnwidth]{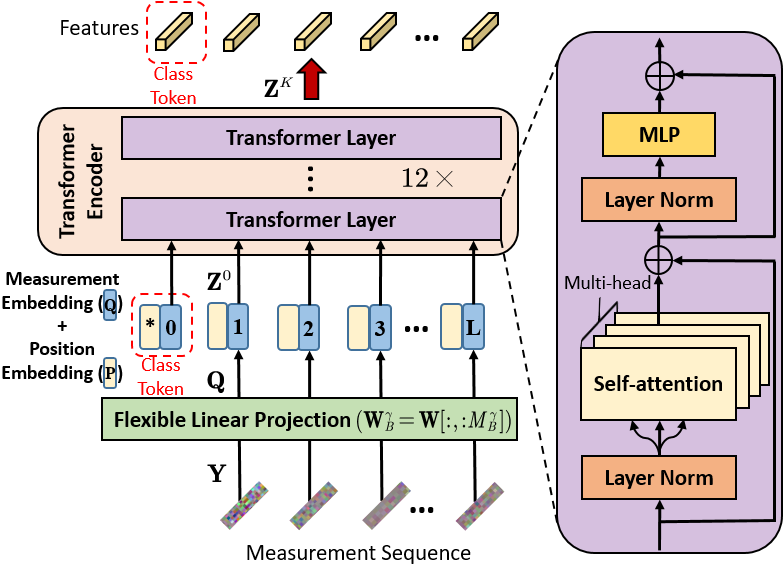}\\(a) Transformer-based Backbone (TB)\\~\\
\end{minipage}
\begin{minipage}{0.38\linewidth}
\centering
\includegraphics[width=1\columnwidth]{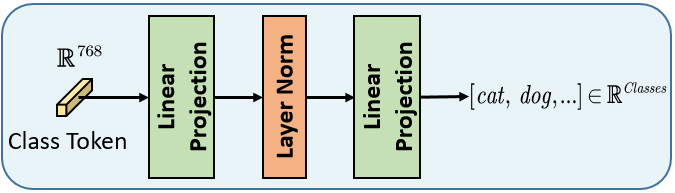}\\(b) Task-oriented Head for Image Classification (THIC)\\~\\\vspace{8pt}
\includegraphics[width=1\columnwidth]{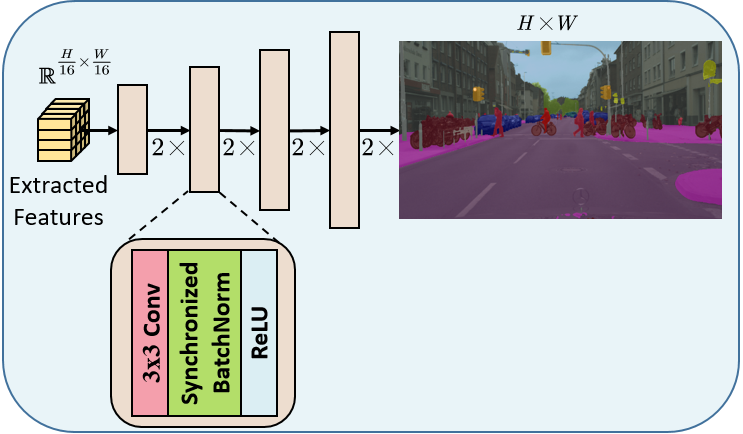}\\(c) Task-oriented Head for Semantic Segmentation (THSS)
\end{minipage}
\caption{Detailed architectures of transformer-based backbone (TB) and task-oriented heads (TH) in our TransCL. (a) presents the details of our transformer-based backbone (TB). (b) is the illustration of task-oriented head for image classification (THIC). (c) presents the details of task-oriented head for semantic segmentation (THSS), which is a plain network to perform upsampling and predicting in a progressive manner.}
\label{network}
\end{figure*}
 
\subsection{Transformer}
Transformer was first introduced in \cite{attis} for machine translation, which is a self-attention mechanism to construct long-range correlations for elements in different places. Given an input sequence $\mathbf{E} \in \mathbb{R}^{C\times L}$ with the sequence length being $L$ and the number of channels being $C$, a pure transformer layer has a global receptive field to build correlations for information transfer. At the beginning of the transformer network, each element of the input sequence is first embedded by a linear projection $\mathbf{W}$$\in$$\mathbb{R}^{d\times C}$ to the length of $d$ and then added by the position embedding $\mathbf{P}$$\in$$\mathbb{R}^{d\times L}$. Let $\mathbf{Z}^l$$\in$$ \mathbb{R}^{d\times L}$ represent the input of the $l$-th transformer layer. The core component is a self-attention module with multiple heads. As illustrated in Fig.~\ref{nl}, each head performs non-local operation in a triplet of embedded features (\textit{query, key, value}) computed from the input as:
\begin{equation}
    query=\mathbf{W}_Q\mathbf{Z}^l,\ key=\mathbf{W}_K\mathbf{Z}^l,\ value=\mathbf{W}_V\mathbf{Z}^l,
\end{equation}
where $\mathbf{W}_Q$,$\mathbf{W}_K$,$ \mathbf{W}_V$$\in$$\mathbb{R}^{d^\prime \times d}$ are the learnable parameters of three linear projection layers. The self-attention operation can be formally defined as:
\begin{equation}
    \text{SA}(\mathbf{Z}^l)=\left(\mathbf{W}_V\mathbf{Z}^l\right)\text{Softmax}\left(\frac{(\mathbf{W}_Q\mathbf{Z}^l)^\top\mathbf{W}_K\mathbf{Z}^l}{\sqrt{d}}\right).
    \label{self}
\end{equation}
However, there exists a gap between $2D$ image and $1D$ sequence in applying transformer to vision tasks. A straightforward method~\cite{nonlocalnet} is to flat an input feature map $\mathbf{F}_{in}$$\in$$\mathbb{R}^{C\times H\times W}$ into a sequence with the length of $HW$. Given the quadratic model complexity of the transformer, it is not possible to handle the input sequence with high dimensions. \cite{vit} proposed a solution to unfold an input image into a sequence of image blocks with each block size being $B$$\times$$ B$, significantly reducing the sequence length to $\frac{HW}{B^2}$. This strategy can reduce the computational complexity of the transformer layer. Since ViT was proposed \cite{vit}, there have been several attempts to adapt transformers towards vision tasks, including image classification \cite{deit,tnt}, object detection \cite{detr,pvt}, segmentation \cite{vtseg}, human pose estimation \cite{vtpos}, and image restoration \cite{cola,vtsr,vtpre}. However, this strategy has hardly been studied in the community of compressive learning (CL) and compressed sensing (CS). 


\section{Methodology}
In this section, we will elaborate on our proposed strong and flexible transformer-based compressive learning framework, dubbed TransCL.

\subsection{Framework}
The global architecture of our TransCL is presented in Fig.~\ref{net_all}, which is mainly composed of a Compressed Sensing Module (CSM), a Transformer-based Backbone (TB), and a Task-oriented Head (TH). Note that our TransCL targets performing high-level vision tasks on real-world images, rectifying the information loss in the measurement domain and handling input with arbitrary CS ratios. CSM is designed to perform sensing and compressing in a block-by-block manner with arbitrary CS ratios, and it is jointly optimized with the whole network. To handle the serialized output of CSM and resist information loss, TB is designed as a pure transformer architecture. A creative modulation is that TB begins with a flexible linear projection (FLP) layer to project the input with arbitrary CS ratios to a set of measurement embeddings with the same vector length. Based on TB, Th utilizes the extracted features to perform specific vision tasks. We will provide the details below.

\subsection{Compressed Sensing Module}
\label{csm}
As illustrated in Fig.~\ref{net_all}, CSM is a data acquisition step via CS, aiming to compress the scene to a small number of measurements through a linear sampling matrix $\mathbf{\Phi}$. In our TransCL, we set $\mathbf{\Phi}$ as a trainable parameter, which is jointly optimized with the whole network. Based on these basic settings, we carry out more elaborate designs to make CSM powerful and flexible.  

\subsubsection{Efficient Block-based Arbitrary Sampling}
\label{sec_arb}
\begin{figure}[t]
    \centering
    \includegraphics[width=.9\columnwidth]{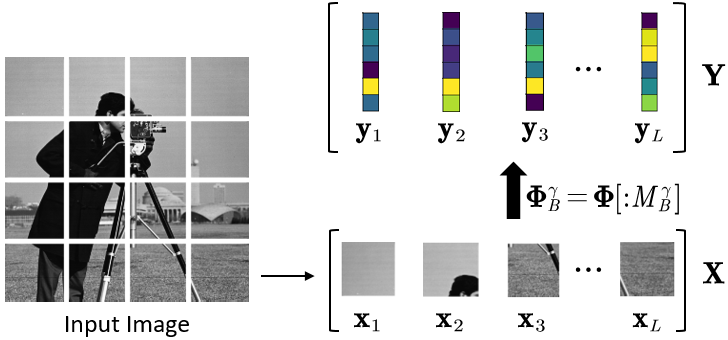}
    \caption{Illustration of compressed sensing module (CSM) in our TransCL.}
    \label{bcs}
\end{figure}

Due to the high computational complexity in sampling large-scale images, most existing CL methods are limited in MNIST/CIFAR-like datasets. To process high-resolution images in real-world applications and provide serialized measurements to the transformer-based backbone (TB), we propose to apply the strategy of block-based compressed sensing (BCS) \cite{bcs}\cite{bcs2} to CL tasks. BCS is able to sense and compress images in a block-by-block manner through the same operator, which has the advantage of decreasing the bandwidth required to transmit measurement vectors between system memory and the DMD array.
Specifically, given a high-resolution input image of size $H$$\times$$ W$. Instead of direct CS by a large sampling matrix $\mathbf{\Phi}$$\in$$\mathbb{R}^{M\times N}$, where $M$=$\gamma N$ and $N$=$HW$. As shown in Fig.~\ref{bcs}, we first divide the image into $L$ non-overlapped blocks with each block size being $B$$\times$$B$ and $L$=$\frac{HW}{B^2}$. Denote $\mathbf{x}_i$$\in$$ \mathbb{R}^{B^2}$ as the vectorized representation of the $i$-th image block and define $\mathbf{X}$$=$$[\mathbf{x}_1,\mathbf{x}_2,...,\mathbf{x}_L]$. The corresponding block-based sampling matrix is represented as $\mathbf{\Phi}^\gamma_B$$\in$$\mathbb{R}^{M_B^\gamma\times B^2}$, where $M_B^\gamma$$=$$\gamma B^{2}$. Then, the resulting CS measurement $\mathbf{y}_i$$\in$$\mathbb{R}^{M_B^\gamma}$ is generated by $\mathbf{y}_i$$=$$\mathbf{\Phi}_B^\gamma\mathbf{x}_i$. 
Finally, our CSM generates a sequence of measurements: $\mathbf{Y}$$=$$[\mathbf{y}_1,\mathbf{y}_2,...,\mathbf{y}_L]$. The output of CSM is formulated as:
\begin{equation}
    \mathbf{Y} =  \mathbf{\Phi}_B^\gamma \mathbf{X}, 
\label{eq:csm}
\end{equation}
where $\mathbf{Y}$$\in$$ \mathbb{R}^{M_B^\gamma \times L}$. Note that BCS only needs to store a block-based sampling matrix instead of a full one and greatly reduces the computational burden of matrix multiplication through a block-by-block strategy, which is quite memory efficient. In addition, BCS can also be easily implemented on hardware by introducing a mask for each square region to drive the DMD of the single-pixel camera to route light from each region to the detector one by one~\cite{hardware_bcs1} \cite{hardware_bcs2}.

\begin{figure}
    \centering
    \begin{minipage}[t]{0.55\linewidth}
\centering
    \includegraphics[width=0.75\columnwidth,height=6cm]{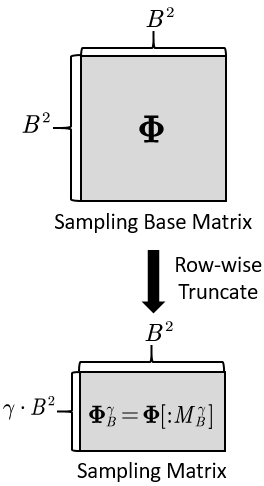}\\ \small (a) Sampling with\\ arbitrary CS ratio
    \end{minipage}
    \begin{minipage}[t]{0.4\linewidth}
\centering
    \includegraphics[width=.9\columnwidth,height=6cm]{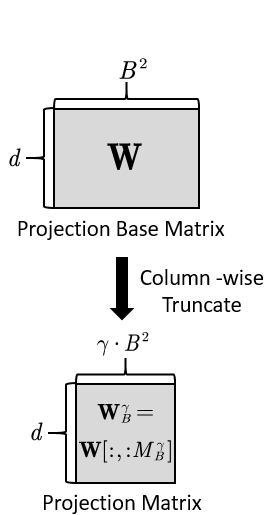}\\ \small (b) Projection with arbitrary CS ratio
    \end{minipage}
    \caption{Illustration of generating the sampling matrix $\mathbf{\Phi}_B^\gamma$ and the projection matrix $\mathbf{W}_B^\gamma$ with arbitrary CS ratio $\gamma$ from their corresponding base matrices $\mathbf{\Phi}$ and $\mathbf{W}$, respectively.}
    \label{mr2}
\end{figure}

Furthermore, unlike existing methods~\cite{op1,op2,cl_w5,mcl_v2,opin} for jointly optimizing the sampling matrix and inference model for a fixed CS ratio each time, we propose a learnable and powerful CSM, which can generate a block-based sampling matrix with arbitrary CS ratio $\gamma$ by training once. Our proposed sampling scheme is presented in Fig.~\ref{mr2}(a). As a reusable scheme, we first define a learnable sampling base matrix (SBM) as $\mathbf{\Phi}$ of size $B^2$$\times$$B^2$. Then, the sampling matrix $\mathbf{\Phi}_B^\gamma$ with any CS ratio $\gamma$ used in Eq.~(\ref{eq:csm}) can be directly constructed by truncating the first $M_B^\gamma$ rows from SMB $\mathbf{\Phi}$, \textit{i.e.}, $\mathbf{\Phi}_B^\gamma$$=$$\mathbf{\Phi}[{:M_B^\gamma]}$. Hence, our CSM only needs one learnable SBM to generate matrices with arbitrary sampling ratios ($B^2$ in total), and it only needs to be trained once, without introducing $B^2$ different sampling matrices as learnable parameters and training for $B^2$ times.

\subsubsection{Hardware-friendly Binary Sampling Design}
Although float-point CS matrices have been widely used, the most efficient and energy-saving hardware utilizes binary sampling matrix ($1$ or $-1$) in real-world applications. This hardware-friendly choice has been studied in the recent reconstruction work~\cite{opin}. In this paper, we exploit this property in the field of CL to further train a binary SBM $\mathbf{\Phi}$ in our CSM. We denote $\widetilde{\mathbf{\Phi}}$ as the binary version of the float-point SBM $\mathbf{\Phi}$, which is computed by $\widetilde{\mathbf{\Phi}}=\mathcal{B}(\mathbf{\Phi})$, where $\mathcal{B}(\cdot)$ is an element-wise operation defined as:
\begin{equation}
    \mathcal{B}(a)=\left\{\begin{array}{ll} 1, & if~~a\geq 0; \\
 -1, & if~~a < 0.
\end{array}\right.
\end{equation}
However, such a setting makes it difficult to calculate the gradient. Inspired by Binarized neural networks (BNN)~\cite{bnn} and differentiable quantization~\cite{bnn_cp}, we introduce a proxy function $\tilde{\mathcal{B}}(\cdot)$ to approximate $\mathcal{B}(\cdot)$, which is defined as:
\begin{equation}
\tilde{\mathcal{B}}\left(a\right)=\left\{\begin{array}{ll} 1, & if~~ a>1; \\
a, & if~~ -1 \leq a \leq 1; \\
-1, & if~~ a<-1.
\end{array}\right.
\end{equation}
Thus, the gradient of $\mathcal{B}(\cdot)$, denoted by  $\mathcal{B}^{\prime}(\cdot)$, can be approximated by the gradient of $\tilde{\mathcal{B}}(\cdot)$, denoted by  $\tilde{\mathcal{B}}^{\prime}(\cdot)$, that is:
\begin{equation}
    \tilde{\mathcal{B}}^{\prime}(a)=\left\{\begin{array}{ll} 1, & if~~ -1 \leq a \leq 1; \\
 0, & otherwise.
\end{array}\right.
\end{equation}
Therefore, $\mathcal{B}(\cdot)$ is used in the forward propagation, while $\tilde{\mathcal{B}}(\cdot)$ is used in back-propagation.

\subsection{Transformer-based Backbone}
Fig.~\ref{network}(a) presents the detailed architecture of our proposed transformer-based backbone (TB), which is used to extract informative features from the serialized measurements $\mathbf{Y}$ with arbitrary CS ratios. TB is composed of a flexible linear projection (FLP) and a stack of transformer layers~\cite{attis} dubbed transformer encoder (TE).

\textbf{Flexible Linear Projection.} As shown in Fig.~\ref{network}(a), our FLP is to project the input measurements $\mathbf{Y}$ of size $M_B^\gamma$$\times$$L$ to the measurement embedding $\mathbf{Q}$ of size $d$$\times$$L$, followed by adding a set of learnable position embedding, denoted by $\mathbf{P}$ of size $d$$\times$$ L$. Note that different sampling matrix $\Phi_B^\gamma$ with different $\gamma$ will produce different $\mathbf{Y}$ with different sizes. To overcome the weakness of existing fixed linear projections that can not handle the measurement vectors with variable lengths and analogous to SBM, we similarly define a trainable projection base matrix (PBM) as $\mathbf{W}$ of size $d$$\times$$B^2$. Then, corresponding to each $\mathbf{\Phi}_B^\gamma$ with CS ratio $\gamma$, we adopt the first $M_B^\gamma$ columns from PBM as the projection matrix $\mathbf{W}_B^{\gamma}$, \textit{i.e.}, $\mathbf{W}_B^{\gamma}=\mathbf{W}[:,:M_B^\gamma]$, which is of size $d$$\times$$M_B^\gamma$. Obviously, we only need to jointly train the PBM $\mathbf{W}$ once, rather than train $B^2$ different projection matrices for $B^2$ times, making our FLP much more flexible. Therefore, as illustrated in Fig.~\ref{network}(a), the output of our FLP layer, denoted by $\mathbf{Z}^0$, can be formulated as:
\begin{equation}
    \mathbf{Z}^0=\mathbf{Q}+\mathbf{P}=\mathbf{W}_B^{\gamma}\mathbf{Y}+\mathbf{P},
    \label{org}
\end{equation}
where $\mathbf{Z}^0$$\in$$ \mathbb{R}^{d \times L}$. Note that an extra class token exists in the image classification task concatenated with $\mathbf{Z}^0$ going through the transformer encoder to collect the category information for prediction. The class token is inherited from \cite{bert} and has been widely used in recent transformer-based image classification researches \cite{vit,deit,tnt}. 

\textbf{Transformer Encoder.} Taking the embedding sequence $\mathbf{Z}^0$ as input, the transformer encoder (TE), constructed by a stack of pure transformer layers~\cite{attis} as shown in Fig.~\ref{network}(a), is deployed to extract deeper features from the CS measurements. The core part of each transformer layer is a multiheaded self-attention (MSA) block. Each head performs the operation defined in Eq.~(\ref{self}), and their outputs are concatenated together. Multilayer perceptrons (MLP), Layer Norm (LN), and residual connections are applied in the tail of each transformer layer. Thus, the output of each transformer layer has the following formulation:
\begin{equation}
\begin{cases}
& \widetilde{\mathbf{Z}}^l=\text{MSA}(\text{LN}(\mathbf{Z}^l)); \\
& \mathbf{Z}^{l+1}=\text{MLP}(\text{LN}(\widetilde{\mathbf{Z}}^l))+\widetilde{\mathbf{Z}}^l,\ l=0,1,2,...,K-1,\\
\end{cases}
\end{equation}
where $K$ is the number of transformer layers (12 by default), and $\mathbf{Z}^{K}$ denotes the output of TE. 

\subsection{Task-oriented Head}
Our TransCL focuses on two high-level vision tasks: image classification and semantic segmentation. For each task, we adopt a specific task-oriented head (TH) to process $\mathbf{Z}^{K}$.

\subsubsection{Task-oriented Head for Image Classification}
As illustrated in Fig.~\ref{network}(b), the take-oriented head for image classification (THIC) is a simple combination of linear projection and layer norm. The input of THIC is the learnable class token (\textit{i.e.}, $\mathbf{Z}^K[:,0]$) with the vector length being $768$, and then the classification head performs a prediction on the top of the class token. Note that the class token only exists in the image classification task. 

\subsubsection{Task-oriented Head for Semantic Segmentation}
The take-oriented head for semantic segmentation (THSS) follows the same architecture as the recent transformer-based method \cite{vtseg}, which takes the original image as input and is also the upper bound model of our TransCL in this task. Since we obtain measurements in an image block of size $16$$\times$$16$, the size of $\mathbf{Z}^K$ is $\frac{W}{16}$$\times$$ \frac{H}{16}$. Through THSS, $\mathbf{Z}^K$ is then gradually upsampled to a segmentation map with the same size as the original image. As illustrated in Fig.~\ref{network}(c), THSS is essentially a simple progressive upsampling model without any bells and whistles. Specifically, each upsampling block is a triple combination of a $1\times 1$ convolutional layer, a Synchronized BatchNorm, and a ReLU~\cite{relu} activation function ending with a bilinear upsampling operation. The simpleness of THSS can also better demonstrate the effectiveness of our proposed TransCL.

\begin{figure}[t]
\centering
\small 
\begin{minipage}[t]{0.24\linewidth}
\centering
\includegraphics[width=1\columnwidth,height=1\columnwidth]{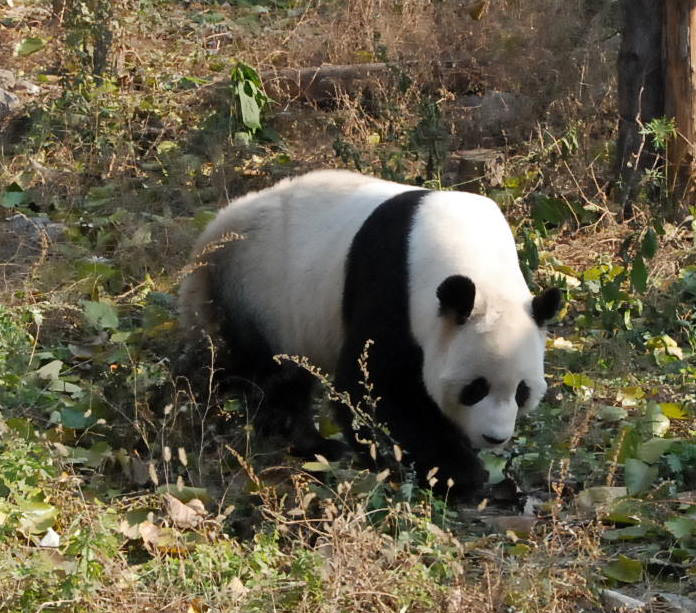}\\Original
\end{minipage}
\begin{minipage}[t]{0.24\linewidth}
\centering
\includegraphics[width=1\columnwidth,height=1\columnwidth]{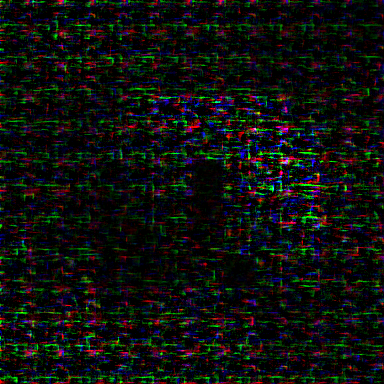}\\Reconstructed
\end{minipage}
\begin{minipage}[t]{0.24\linewidth}
\centering
\includegraphics[width=1\columnwidth,height=1\columnwidth]{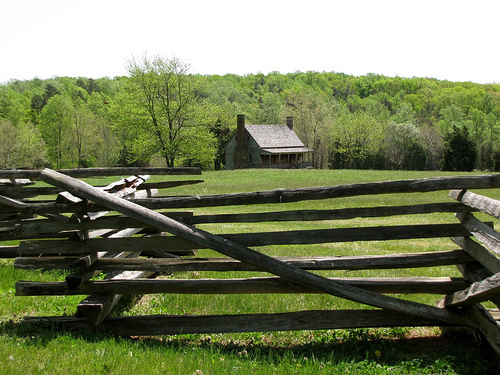}\\Original
\end{minipage}
\begin{minipage}[t]{0.24\linewidth}
\centering
\includegraphics[width=1\columnwidth,height=1\columnwidth]{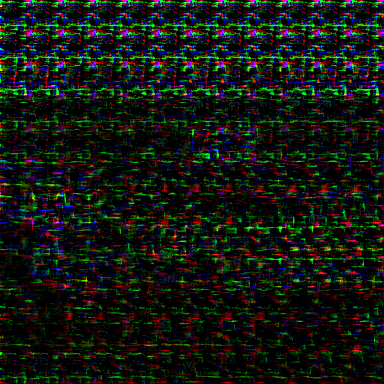}\\Reconstructed
\end{minipage}
\centering
\caption{Visualization of privacy protection. We present the original image and reconstructed result generated by~\cite{opin} without knowing the sampling matrix.}
\label{ab_pri} 
\end{figure}

\subsection{Privacy Protection}
It is worth emphasizing that our TransCL is superior in privacy protection. On the one hand, the CS measurements have little visual information, which is an inherent property of CS. Although there exist several CS reconstruction methods, \textit{e.g.}, \cite{cs_w1,cs_w2,opin,ista}, they all need to know the corresponding sampling matrix, especially at low CS ratios. To verify it, we adopt the recent~\cite{opin} for reconstruction with the CS measurements generated by our trained CSM rather than its own learned sampling matrix. The reconstructed results are presented in Fig.~\ref{ab_pri}, from which one can observe that it is almost inaccessible to original images without the corresponding sampling matrix. On the other hand, the transformer-based backbone (TB) in our TransCL can further strengthen the ability of privacy protection due to the serialized intermediate features, which are more abstract and invisible than CNN counterparts. Thus, we believe that our TransCL can protect data privacy in the process of data transmission and processing.

\section{Experiment}
To demonstrate the superiority of our proposed TransCL, we apply it to two representative high-level vision tasks, \textit{i.e.}, image classification and semantic segmentation. The experiment design comes from two factors. On the one hand, studying several linear projection layers under traditional problem settings with tiny images has little significance for promoting CL. The urgent problem faced by CL is how to handle real-world benchmarks and perform inference as flexibly as image-domain methods. On the other hand, though our approach is a new formulation of CL under new problem settings, only comparing our TransCL with image-domain methods on real-world benchmarks is inadequate. We also need to compare our TransCL with existing CL methods (\textit{i.e.}, VCL~\cite{vcl} and MCL~\cite{cl_w5}), which are difficult to handle large-scale images. Thus, in the experiment, we focus on comparing our TransCL with the recent image-domain methods to present that CL can also be applied to complex and real-world benchmarks (\textit{e.g.}, ImageNet~\cite{imagenet}, ADE20K~\cite{ade20k}, PASCAL Context \cite{voc}, and Cityscapes \cite{city}) without rigid restrictions from data scale and CS ratio. Additionally, we also train and evaluate our TransCL on tiny datasets (\textit{e.g.}, CIFAR10 and CIFAR100~\cite{cifar}) to directly compare with existing CL methods. 

\subsection{Model Details and Variants}
In image classification, the transformer architecture is the same as the base vision transformer (ViT-B), which contains $12$ transformer layers, with the number of heads being $12$ in each layer. The dimension of the measurement embedding $\mathbf{Q}$ is set as $768$. In semantic segmentation, we utilize the same backbone (ViT-L) as the upper-bound model (SETR). It contains $24$ transformer layers, with the number of heads being $16$. The dimension of the measurement embedding $\mathbf{Q}$ is set as $1024$. Several variants of our proposed TransCL are defined with different names. For instance, ``TransCL-16-10'' represents the variant with the block size of $16$$\times$$16$ and CS ratio of $10\%$. In our experiments, we evaluate our TransCL with both fixed CS ratios and arbitrary CS ratios. The fixed CS ratios include $10\%$, $5\%$, $2.5\%$, and $1\%$. We denote $arb$ as the suffix of the variant that can handle input measurements with arbitrary CS ratios. The suffixes of block size include $16$ and $32$. 

\subsection{Image Classification}
\textbf{Experiment Settings.} Image classification is the most representative high-level vision task. For this application, we apply our model to the ImageNet 2012~\cite{imagenet}, CIFAR-10, and CIFAR-100~\cite{cifar} datasets:
\begin{itemize}
    \item \textbf{ImageNet 2012} contains $1.28$ million training images and $50,000$ validation images from $1,000$ categories. The image resolution is set as $384$$\times$$384$. It is a commonly used benchmark in the image domain. However, limited by the computational cost, little CL method can be applied to this dataset at such scale.
    \item \textbf{CIFAR-10} consists of $60,000$ $32$$\times$$ 32$ colour images in $10$ classes, with $6,000$ images per class. There are $50,000$ training images and $10,000$ test images. 
    \item \textbf{CIFAR-100} is just like the CIFAR-10, except it has $100$ classes. For each class, there are $500$ training images and $100$ testing images.
\end{itemize}
Note that CIFAR-10 and CIFAR-100 are two datasets with tiny images. The transformer architecture has difficulty in processing tiny input. ViT~\cite{vit} upsampled images from these two datasets and then performed inference. Considering the upsampling will introduce additional information into CS measurements, we interpolate the measurements $\mathbf{Y}$ ($\mathbb{R}^{M_B^\gamma\times L^\prime}$$\mapsto $$\mathbb{R}^{M_B^\gamma \times L}$) instead of the input images. In this way, no extra information is involved in the measurements. During training, we apply the same training settings as the upper bound model (ViT-B~\cite{vit}) of our TransCL. Specifically, we use SGD with momentum being $0.9$ and the weight-decay being $10^{-4}$ as the optimizer. The batch size is set as $256$, and we train our proposed TransCL on 4 NVIDIA V100 GPUs. Realizing the importance of the pre-trained model in transformer-based methods, we initialize the parameters in TB with the pre-trained model of ViT-B. The CSM, TB, and TH in TransCL are jointly optimized by a cross-entropy loss. Moreover, traditional ViT requires large-scale datasets (\textit{e.g.}, ImageNet-21K) for pre-training. Recently, some strategies were proposed to build data-efficient ViT. For instance, the timm~\cite{timm} presents a training procedure to improve the top-1 accuracy of ViT-B from 77.91$\%$ to 79.35$\%$ with ImageNet-1K only. The recent DeiT~\cite{deit} utilizes a more elaborate optimization process and knowledge distillation strategies to further improve the performance. Inspired by DeiT, in this paper, we also present a data-efficient version (\textit{i.e.}, \alambic ), which is trained with knowledge distillation on ImageNet-1K only. The distillation process and teacher network follow the same design as DeiT.

\begin{table}[h]
    \centering
        \caption{Image classification performance on validation dataset of ImageNet. “Top-1 ACC” denotes the top-1 ($\%$) accuracy. Performances of different methods, model parameters, and the number of measurements for each image are reported.
        }
        \footnotesize
    \begin{tabular}{l|l|c|c}
    \bottomrule
         Method & \ Meas. & Param./Flops& Top-1 ACC \\
        \bottomrule
         ResNet-18\cite{resnet} & 3$\times$50176 & \textbf{12}M/\textbf{1.9}G & 69.83\\
         ResNet-50\cite{resnet} & 3$\times$50176 &25M/4.2G &76.20\\
         ResNet-101\cite{resnet} & 3$\times$50176 & 45M/7.8G & 77.41\\
         ResNet-152\cite{resnet} & 3$\times$50176 & 60M/11.58G & 78.33\\
         ViT-B-16\cite{vit} & 3$\times$147456 & 86M/49.3G & \textbf{83.97}\\
         ViT-B-32\cite{vit} & 3$\times$147456 & 88M/12.3G& 81.28\\
         PVT-Large\cite{pvt} & 3$\times$50176 & 62M/9.9G & 81.71\\
         DeiT-B $\uparrow384$\cite{deit} & 3$\times$147456 & 86M/49.4G & 83.12\\
         Twins\cite{twins} & 3$\times$50176 & 99M/14.8G & 83.70\\
         \hline
         VCL-T-10 & 3$\times$14745 & 589M/50.8 & 67.59\\
         VCL-T-1 & 3$\times$1474 & 136M/49.6 & 59.89\\
         MCL-T-10 & 3$\times$14770 & 87M/49.4 & 74.07\\
         MCL-T-1 & 3$\times$1477 & 86M/49.3 & 68.92\\
         \hline
         TransCL-16,32-10 & 3$\times$14745 & 86,88M/49.3,12.3G & 83.86,81.82\\
         TransCL-16,32-5 & 3$\times$7372 & 86,88M/49.3,12.3G & 83.05,81.53\\
         TransCL-16,32-2.5 & 3$\times$3686 & 86,88M/49.3,12.3G & 81.34,80.50\\
         TransCL-16,32-1 & \textbf{3$\times$1474} & 86,88M/49.3,12.3G & 78.86,78.00\\
         \hline
         TransCL-16-10$\alambic$ & 3$\times$14745 & 86M/49.3G & 83.29\\
         TransCL-16-5$\alambic$ & 3$\times$7372 & 86M/49.3G & 82.46\\
         TransCL-16-2.5$\alambic$ & 3$\times$3686 & 86M/49.3G & 81.24\\
         TransCL-16-1$\alambic$ & \textbf{3$\times$1474} & 86M/49.3G & 78.58\\
    \hline
    \end{tabular}
    \label{class}
\end{table}

\begin{table}[h]
    \centering
        \caption{Image classification performance on validation datasets of CIFAR-10 and CIFAR-100~\cite{cifar}. Top-1 ($\%$) accuracy is presented below. 
        }
    \footnotesize
    \begin{tabular}{c|l|l|l|c|c}
    \bottomrule
         \makecell[c]{CS\\ Ratio} & Method & Backbone & Meas. &  \makecell[c]{CIFAR\\-10} &  \makecell[c]{CIFAR\\-100} \\
        \bottomrule
         \multirow{6}*{$25\%$} & VCL~\cite{vcl} & ResNet110 & 768 & 78.56 & 53.03\\
         & MCL~\cite{cl_w5} & ResNet110 & 760 & 89.22 & 67.21\\
         & VCL-T & ViT-B-32 & 768 & 93.80 & 78.65\\
         & MCL-T & ViT-B-32 & 760 & 94.51 & 78.60\\
         & TransCL-16 & ViT-B-16 & 760 & 94.35 & 81.45\\
         & TransCL-32 & ViT-B-32 & 760 & \textbf{95.18} & \textbf{82.33}\\
        \hline
        \multirow{6}*{$10\%$} & VCL~\cite{vcl} & ResNet110 & 306 & 67.65 & 47.90\\
        & MCL~\cite{cl_w5} & ResNet110 & 308 & 84.74 & 60.30\\
         & MCLwP~\cite{mcl_v2} & AllCNN~\cite{AIICNN} & 308 & 85.84 & 59.83\\
         & VCL-T & ViT-B-32 & 306 & 89.22 & 70.22\\
         & MCL-T & ViT-B-32 & 308 & 90.16 & 67.53\\
        & TransCL-16 & ViT-B-16 & 306 & 90.66 & \textbf{72.80}\\
         & TransCL-32 & ViT-B-32 & 306 & \textbf{91.57} &72.11\\
         \hline
         \multirow{6}*{$1.8\%$} & VCL~\cite{vcl} & ResNet110 & 54 & 61.96 & 41.03\\
         & MCL~\cite{cl_w5} & ResNet110 & 54 & 64.14 & 33.67\\
         & VCL-T & ViT-B-32 & 54 & 61.96 & 41.86\\
         & MCL-T & ViT-B-32 & 54 & 62.28 & 36.43\\
         & TransCL-16 & ViT-B-16 & 54 & 68.44 & 42.66\\
         & TransCL-32 & ViT-B-32 & 54 & \textbf{69.60} & \textbf{42.75}\\
    \hline
    \end{tabular}
    \label{class_cifar}
\end{table}

\begin{figure}[h]
\centering
\small 
\includegraphics[width=.95\columnwidth]{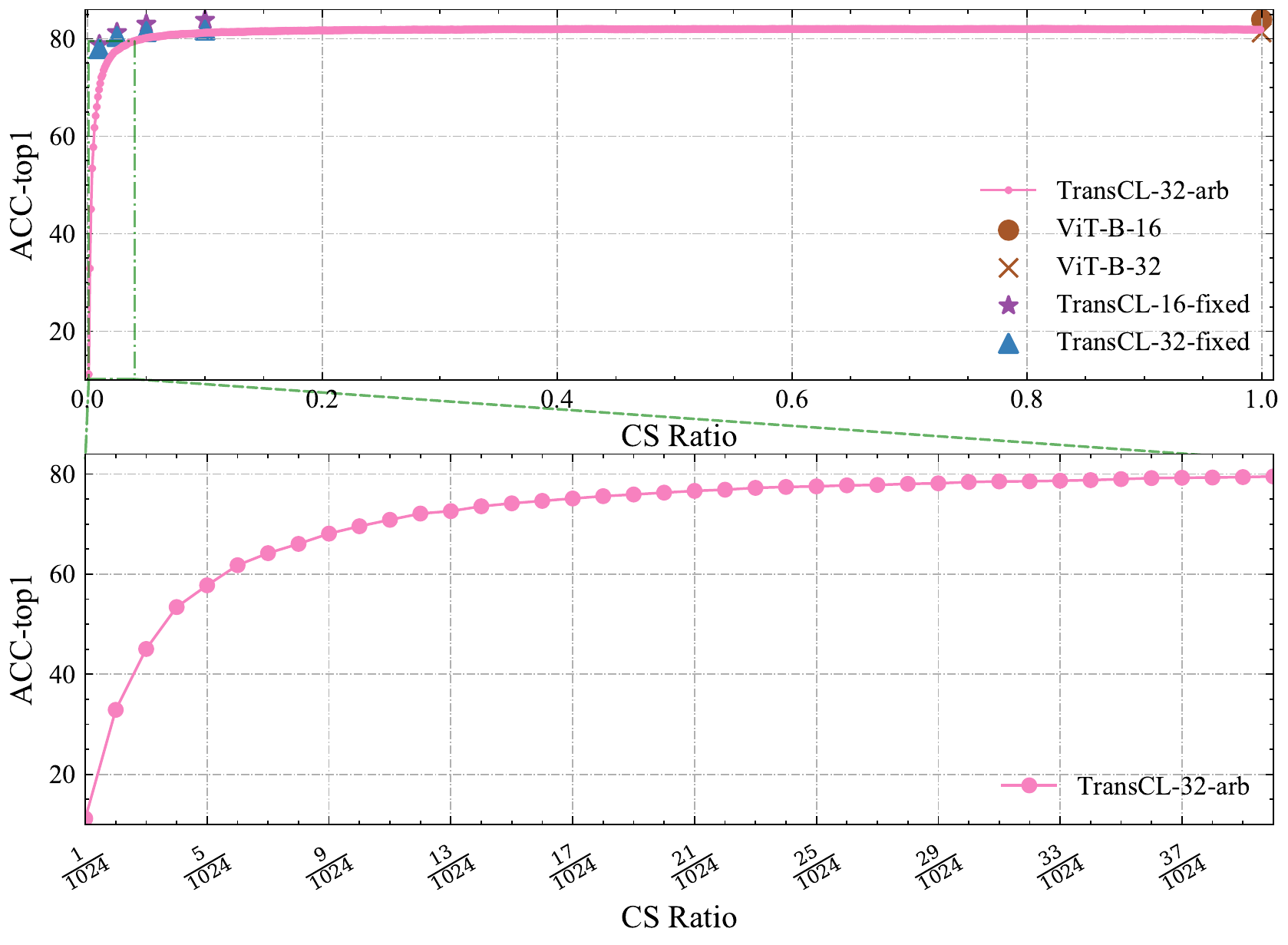}
\centering
\caption{Image classification performance of a single model (TransCL-32-arb) for handling input with arbitrary CS ratios. The first row presents the performance comparison in the case of input with a fixed CS ratio and arbitrary CS ratios. The second row is a partial enlargement of the region labeled with a green box in the first row, presenting the high robustness of our method in handling extremely low and arbitrary CS ratios.}
\label{ab} 
\end{figure}

\begin{figure*}[h]
\centering
\begin{minipage}[t]{0.01\linewidth}
\centering
~\\
\makebox[\hh]{\rotatebox[origin=l]{90}{\makebox[\h][c]{\hspace{-\h}\normalsize{Cityscapes\ \ \ \ \ \ \ PASCAL}}}}\\~\\~\\~\\~\\~\\~\\~\\ \vspace{0.3cm}
\makebox[\hh]{\rotatebox[origin=l]{90}{\makebox[\h][c]{\hspace{-\h}\normalsize{ADE20K}\ \ }}}\\
\end{minipage}
\begin{minipage}[t]{0.19\linewidth}
\centering
\includegraphics[width=1\columnwidth]{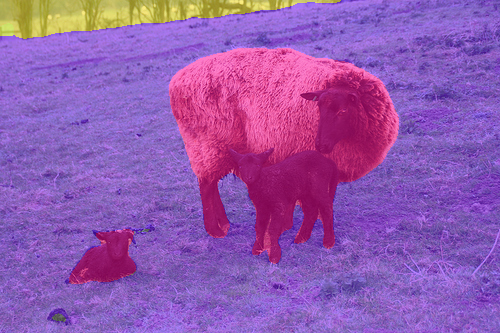}\\
\includegraphics[width=1\columnwidth,height=.6\columnwidth]{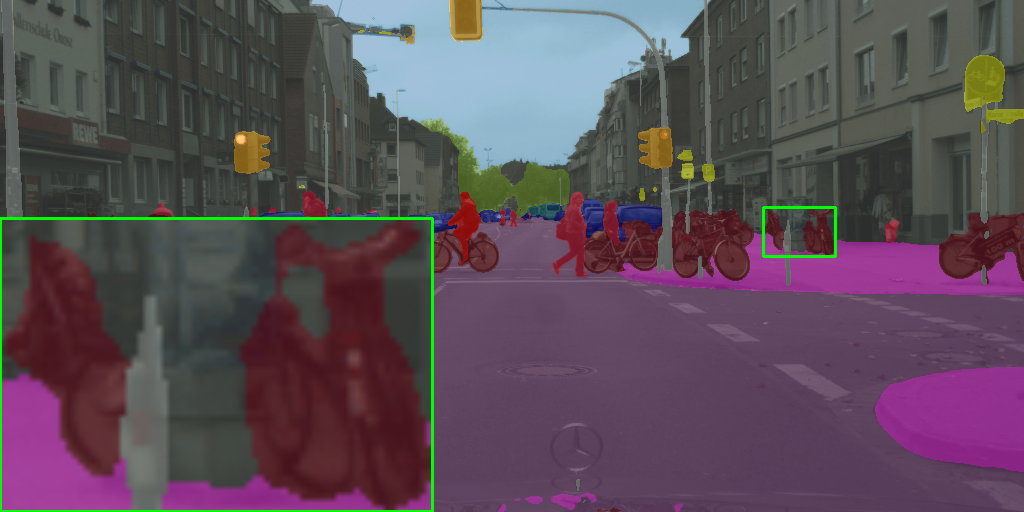}\\
\includegraphics[width=1\columnwidth]{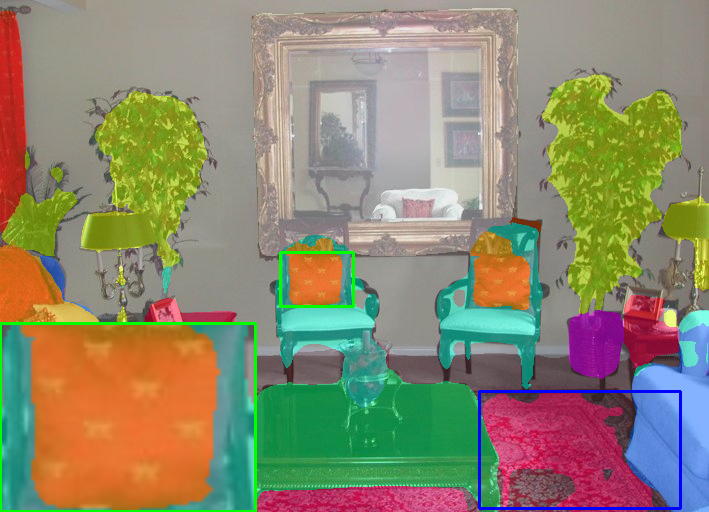}\\SETR~\cite{vtseg}
\end{minipage}
\begin{minipage}[t]{0.19\linewidth}
\centering
\includegraphics[width=1\columnwidth]{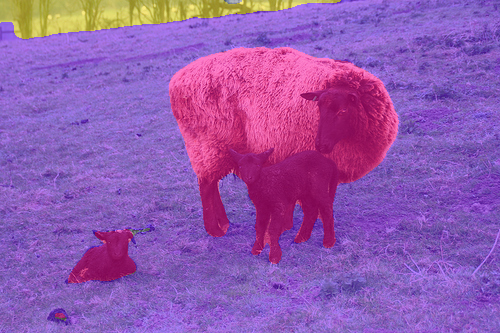}\\
\includegraphics[width=1\columnwidth,height=.6\columnwidth]{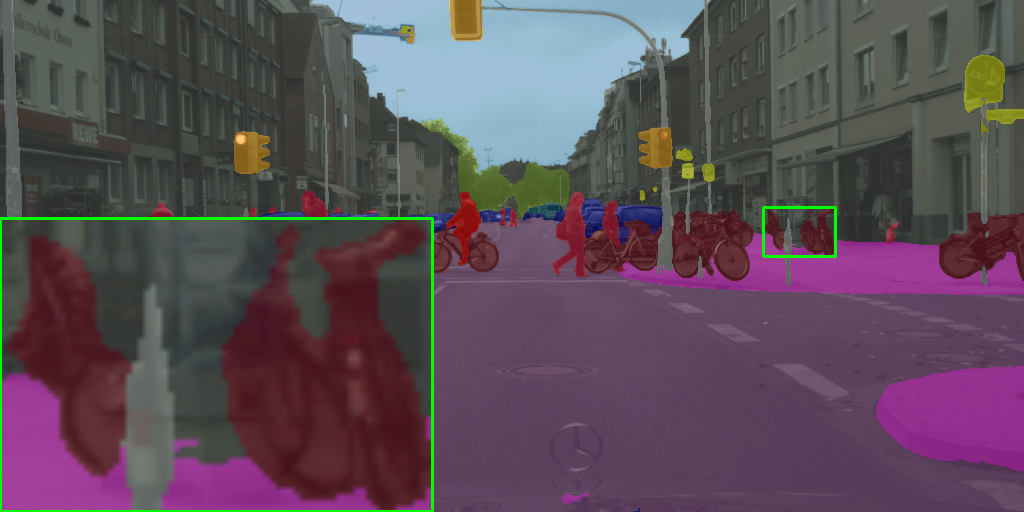}\\
\includegraphics[width=1\columnwidth]{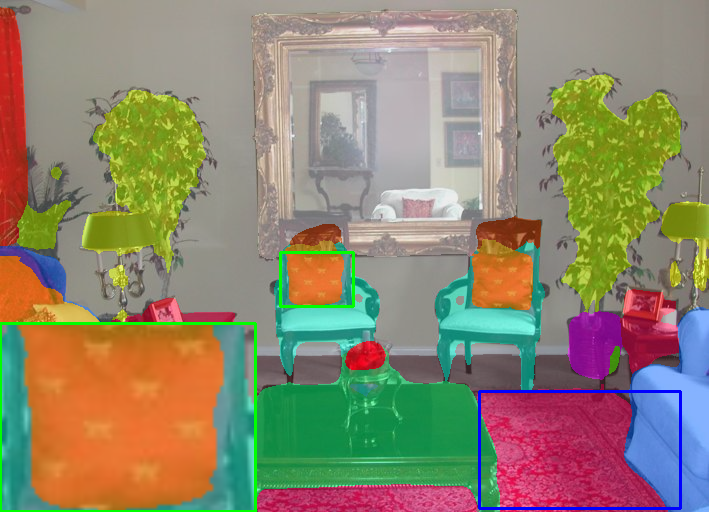}\\TransCL-16-10
\end{minipage}
\begin{minipage}[t]{0.19\linewidth}
\centering
\includegraphics[width=1\columnwidth]{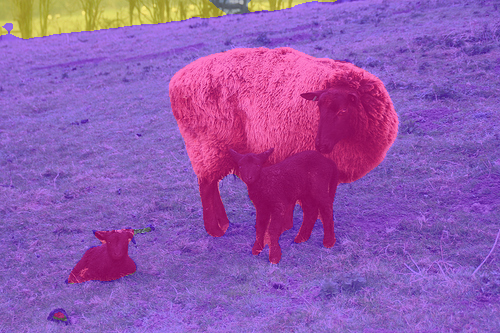}\\
\includegraphics[width=1\columnwidth,height=.6\columnwidth]{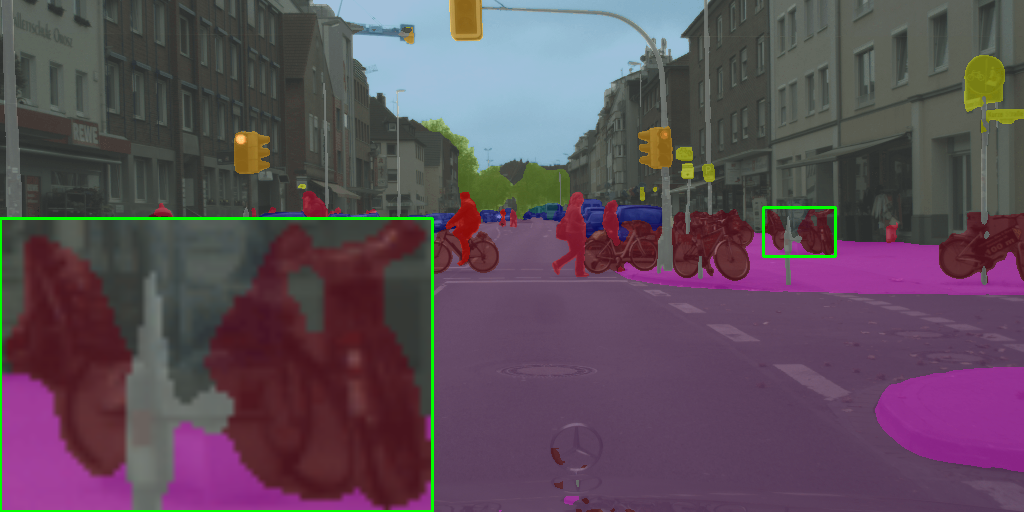}\\
\includegraphics[width=1\columnwidth]{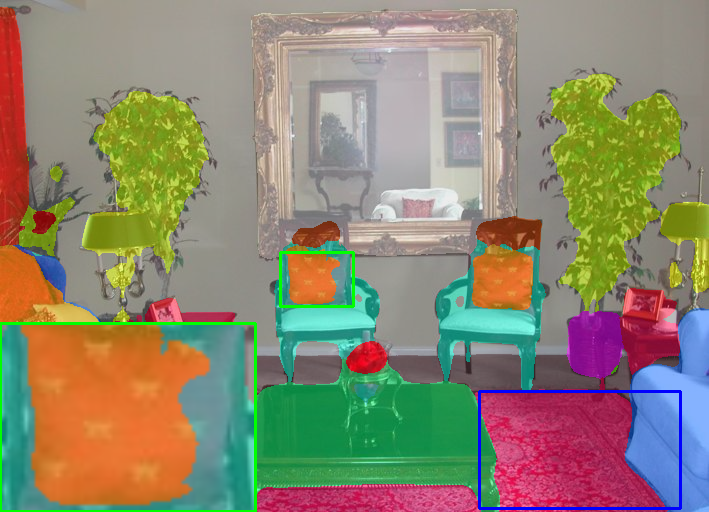}\\TransCL-16-5
\end{minipage}
\begin{minipage}[t]{0.19\linewidth}
\centering
\includegraphics[width=1\columnwidth]{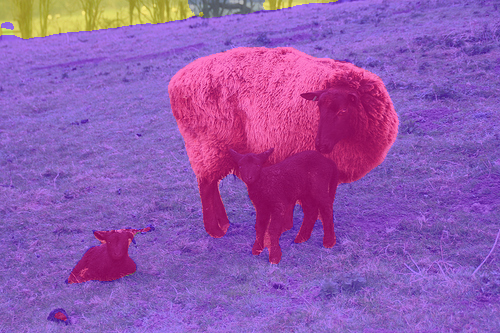}\\
\includegraphics[width=1\columnwidth,height=.6\columnwidth]{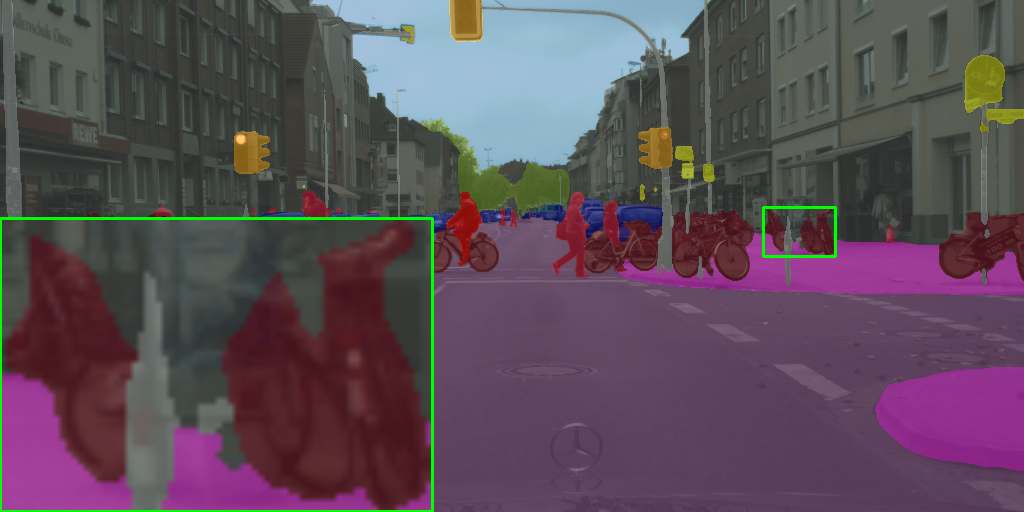}\\
\includegraphics[width=1\columnwidth]{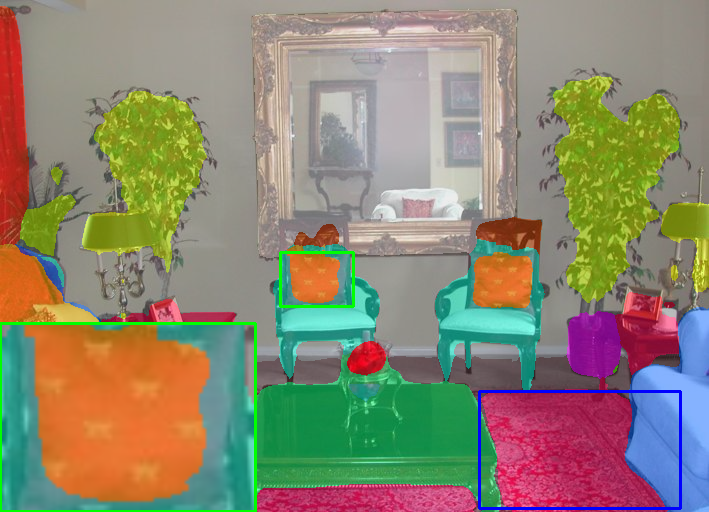}\\TransCL-16-2.5
\end{minipage}
\begin{minipage}[t]{0.19\linewidth}
\centering
\includegraphics[width=1\columnwidth]{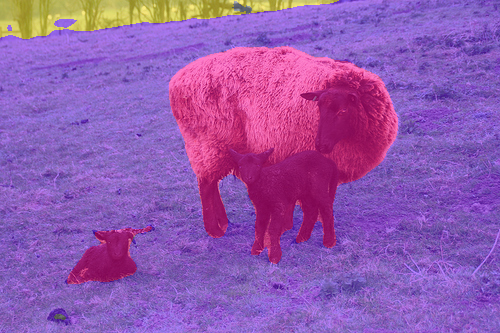}\\
\includegraphics[width=1\columnwidth,height=.6\columnwidth]{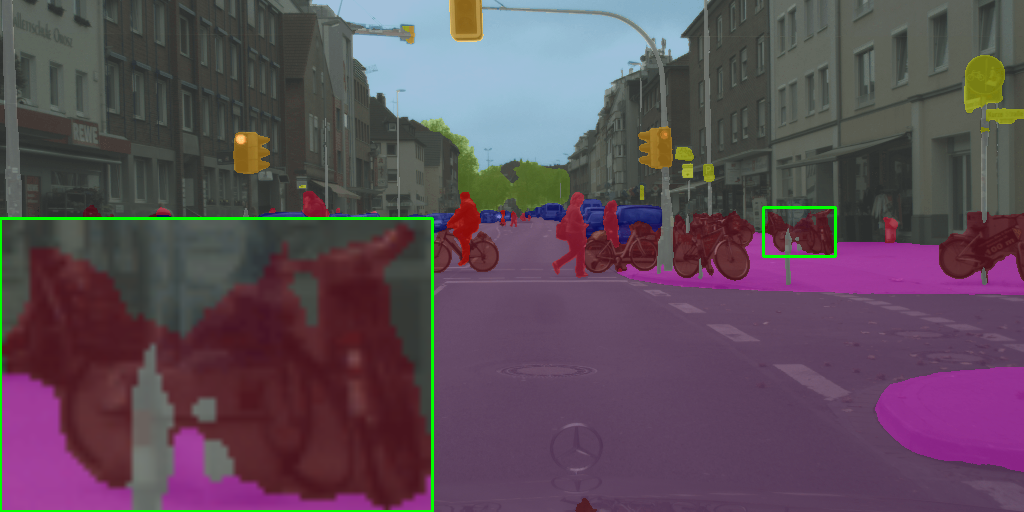}\\
\includegraphics[width=1\columnwidth]{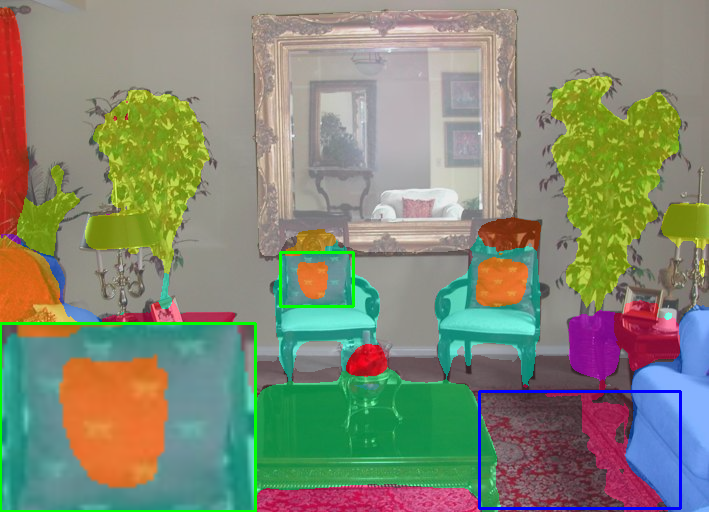}\\TransCL-16-1
\end{minipage}
\centering
\caption{Visual comparison of semantic segmentation. We visualize the segmentation result of samples from PASCAL Context \cite{voc}, Cityscapes \cite{city}, and ADE20K \cite{ade20k}, which are presented in the first row, the second row, and the third row, respectively.}
\label{im_seg} 
\end{figure*}

\begin{figure*}[h]
\centering
\footnotesize
\begin{minipage}[t]{0.33\linewidth}
\centering
\includegraphics[width=1\columnwidth]{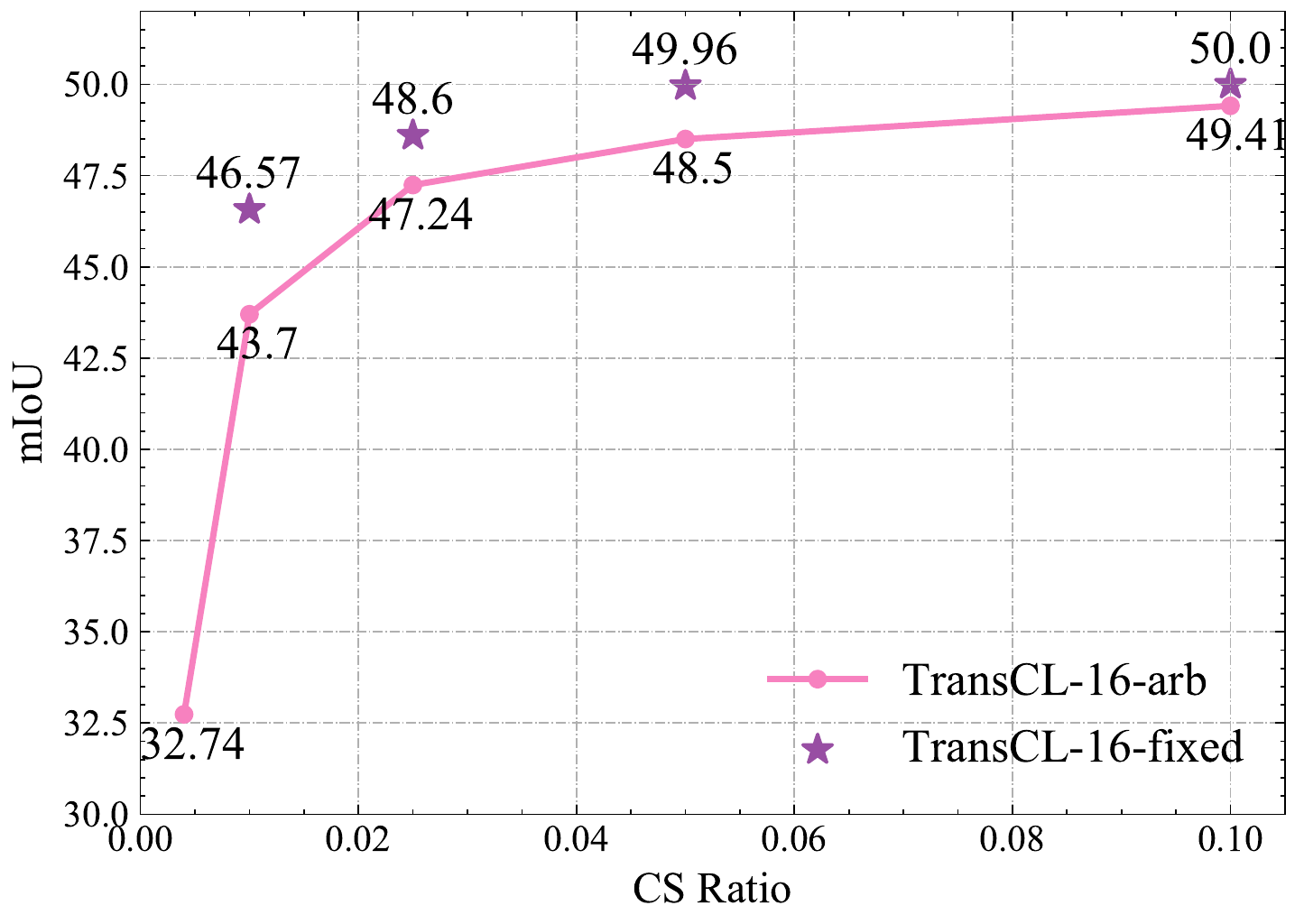}\\\footnotesize{ADE20K~\cite{ade20k}}\\
\end{minipage}
\begin{minipage}[t]{0.33\linewidth}
\centering
\includegraphics[width=1\columnwidth]{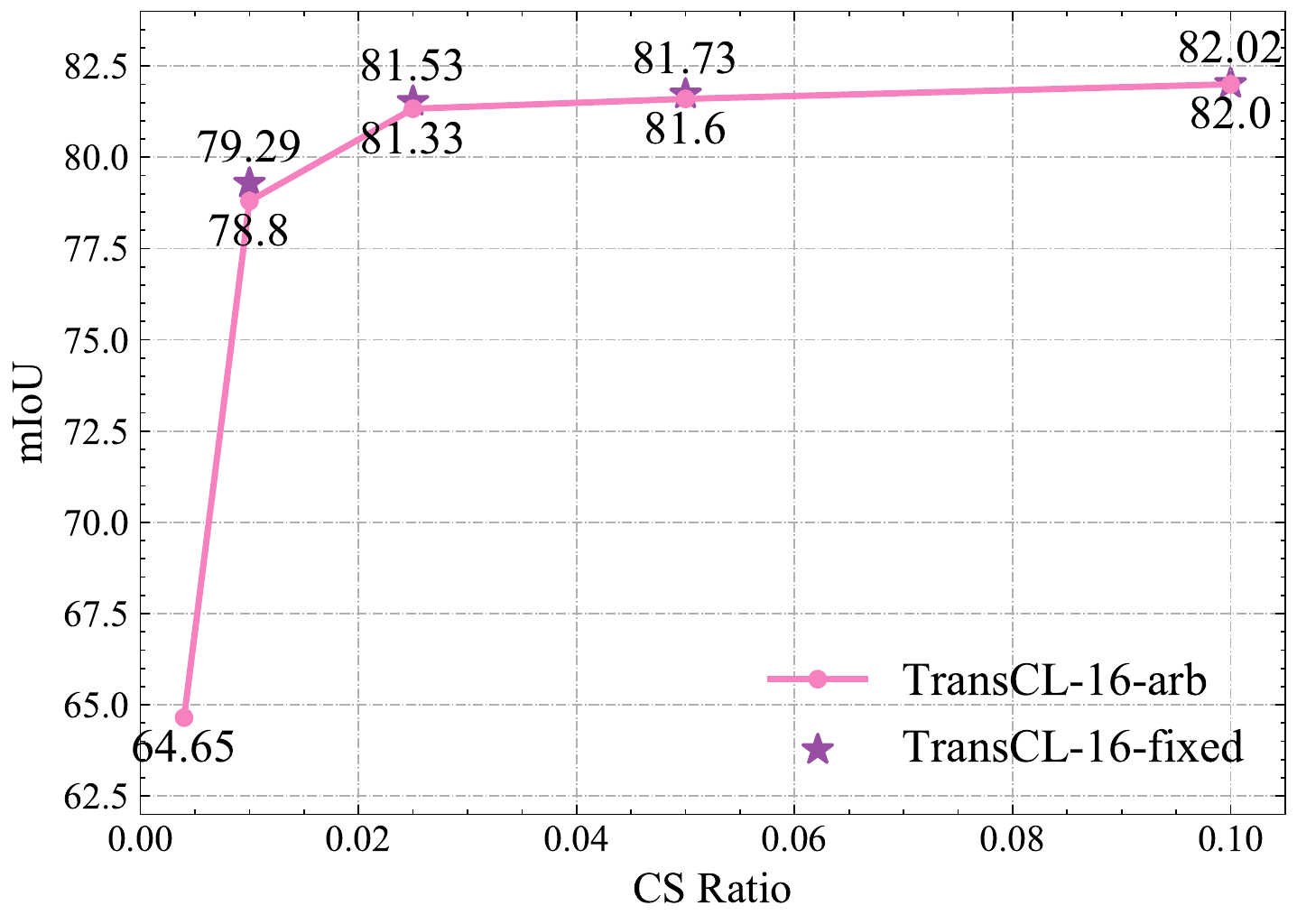}\\\small Cityscapes~\cite{city}\\
\end{minipage}
\begin{minipage}[t]{0.33\linewidth}
\centering
\includegraphics[width=1\columnwidth]{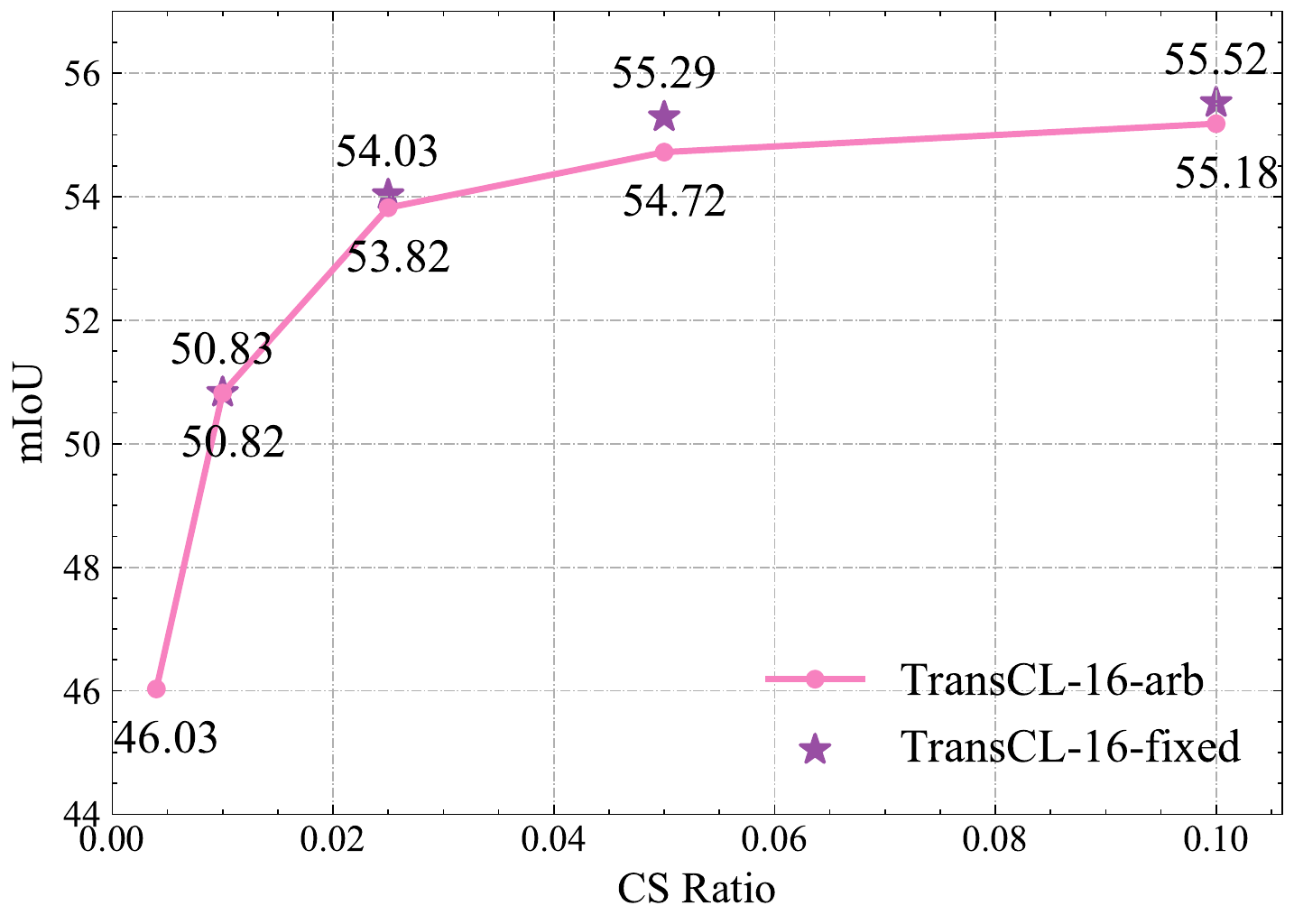}\\\small PASCAL Context~\cite{voc}\\
\end{minipage}
\centering
\caption{Semantic segmentation performance of our TransCL with arbitrary CS ratios on three benchmarks.}
\label{ab_seg}
\end{figure*}

\textbf{Comparison Results.} The evaluation is conducted on the large-scale ImageNet and tiny CIFAR-10/100 datasets. We compare our TransCL with some image-domain methods, \textit{e.g.}, the well-known ResNet~\cite{resnet} and transformer-based methods~\cite{vit,pvt,deit,twins}. We also compare our method with existing CL methods, \textit{i.e.}, VCL~\cite{vcl} and MCL~\cite{cl_w5,mcl_v2}. To fully demonstrate the superiority of our method, we also replace the CNN-based backbone in VCL and MCL with ViT-B model, dubbed VCL-T and MCL-T, respectively. We apply the same training settings to optimize these two models. We report the comparison results of different methods on the ImageNet validation set in Tab.~1. One can see that compared with image-domain methods, our proposed TransCL achieves state-of-the-art performance while utilizing significantly fewer measurements. Even if only $1\%$ measurements are available, our TransCL can still achieve $78.86\%$ classification accuracy. Furthermore, our data-efficient version (\textit{i.e.}, $\alambic$) can achieve comparable performance, trained on the ImageNet-1K only. VCL-T and MCL-T can not perform well on such large-scale benchmarks, and they produce a lot of extra computational complexity. More details can be found in Sec.~5.3. The evaluation on tiny CIFAR-10/100 datasets is mainly used to compare our method with existing CL methods. Note that the measurement results of VCL-T and MCL-T are tensors with the same size ($32\times 32$) as the input, which is not suitable as input of ViT. We utilize bilinear upsampling to resize them to $384\times 384$ and then feed them to the ViT backbone. The results are presented in Tab.~2, presenting that our TransCL significantly outperforms recent top-performing CL methods~\cite{cl_w5,mcl_v2}. In addition, as the number of categories increases or the CS ratio decreases, the margin in performance gains of our proposed TransCL becomes greater. The results in Tab.~1 and Tab.~2 demonstrate that our proposed TransCL performs well on high-resolution images with large visual redundancy and tiny images with intensive information.

\textbf{Performance with arbitrary CS ratios.} Handling measurements with arbitrary CS ratios is valuable and practical, but it has not been explored in existing CL tasks. In this part, we study image classification performance in the case of input measurements with arbitrary CS ratios. The classification performance is presented in Fig.~\ref{ab}. We can find that our proposed TransCL can handle $1,024$ different CS ratios ($\gamma\in [\frac{1}{1024},1]$) with a single model while maintaining stable performance. The enlarged local part in the second row of Fig.~\ref{ab} shows that the performance of our TransCL rapidly reaches the peak with the increase of CS ratio and then becomes stable. In the extreme case where the CS ratio is $\frac{1}{1024}$ (only about $148$ measurements are available for each $384$$\times$$ 384$ image), our TransCL can still achieve $12\%$ classification accuracy of $1,000$ categories. In addition, compared with the variants at fixed CS ratios, the performance loss is minimal, presenting the strong robustness of our TransCL.

\begin{table}[h]
\caption{Quantitative comparison of semantic segmentation task on ADE20K~\cite{ade20k} dataset. Performances of different methods and the number of measurements for each image are reported.
}
\centering
\footnotesize
\begin{tabular}{l|l|c|l}
\bottomrule
Methods (Backbone) & Meas. & Param./Flops& mIoU/Acc\\
\bottomrule
PSP\cite{pspnet} (ResNet-269) & $3$$\times$$223729$ & \textbf{73}M/\textbf{246}G & 44.94/81.69\\
GFF\cite{gffnet} (ResNet-101) & $3$$\times$$262144$ & 141M/811G & 45.33/-\\
APC\cite{apcnet} (ResNet-101) & $3$$\times$$331776$ & 76M/357G & 45.38/-\\
Twins\cite{twins} (SVT-L) & $3$$\times$$262144$ & 133M/297G & 48.80/-\\
SETR\cite{vtseg} (ViT-L-16) & $3$$\times$$262144$ & 309M/316G & \textbf{50.28}/83.46\\
\hline
Ours (TransCL-16-10) & $3$$\times$$26214$ &309M/316G & 50.00/\textbf{83.51}\\
Ours (TransCL-16-5) & $3$$\times$$13107$ &309M/316G & 49.96/83.44\\
Ours (TransCL-16-2.5) & $3$$\times$$6553$ &309M/316G & 48.60/83.13\\
Ours (TransCL-16-1) & \textbf{$\mathbf{3}$$\mathbf{\times}$$\mathbf{2621}$} & 309M/316G & 46.57/82.14\\
\hline
\end{tabular}
\label{seg_ade}
\end{table}

\begin{table}[h]
\caption{Quantitative comparison of semantic segmentation task on Cityscapes~\cite{city} dataset. Performances of different methods and the number of measurements for each image are reported.
}
\centering
\footnotesize
\begin{tabular}{l|l|c|c}
\bottomrule
Methods (Backbone) & Meas. & Param./Flops & mIoU\\
\bottomrule
PSP \cite{pspnet} (ResNet-101) & 3$\times$508369 & 73M/\textbf{554}G & 78.50\\
CCNet \cite{ccnet} (ResNet-101) & 3$\times$591361 & \textbf{71}M/698G & 80.20 \\
GFF \cite{gffnet} (ResNet-101) & 3$\times$746496 & 141M/2305G & 80.40\\
SETR \cite{vtseg} (ViT-L-16) & 3$\times$589825 & 309M/818G & \textbf{82.15}\\
\hline
Ours (TransCL-16-10) & 3$\times$58982 & 309M/818G & 82.02\\
Ours (TransCL-16-5) & 3$\times$29491 & 309M/818G & 81.73\\
Ours (TransCL-16-2.5) & 3$\times$14745 & 309M/818G & 81.53\\
Ours (TransCL-16-1) & \textbf{3$\times$5898} & 309M/818G & 79.29\\
\hline
\end{tabular}
\label{seg_city}
\end{table}

\begin{table}[h]
\caption{Quantitative comparison of semantic segmentation task on Pascal Context~\cite{voc} dataset. Performances of different methods and the number of measurements for each image are reported.
}
\centering
\footnotesize
\begin{tabular}{l|l|c|c}
\bottomrule
Methods (Backbone) & Meas. & Param./Flops & mIoU\\
\bottomrule
PSP \cite{pspnet} (ResNet-101) & 3$\times$223729 & \textbf{73}M/\textbf{246}G & 47.80\\
GFF \cite{gffnet} (ResNet-101) & 3$\times$262144 & 141M/811G & 54.20\\
APC \cite{apcnet} (ResNet-101) & 3$\times$262144 & 76M/282G & 54.70\\
SETR \cite{vtseg} (ViT-L-16) & 3$\times$230400 & 309M/281G & \textbf{55.83}\\
\hline
Ours (TransCL-16-10) & 3$\times$23040& 309M/281G & 55.52\\
Ours (TransCL-16-5) & 3$\times$11520& 309M/281G & 55.29\\
Ours (TransCL-16-2.5) & 3$\times$5760& 309M/281G & 54.03\\
Ours (TransCL-16-1) & \textbf{3$\times$2304}& 309M/281G & 51.83\\
\hline
\end{tabular}
\label{seg_voc}
\end{table}

\subsection{Semantic Segmentation}
\textbf{Experiment Settings.} For the application of semantic segmentation, we conduct experiments on the following three widely-used benchmarks:
\begin{itemize}
    \item \textbf{Cityscapes} \cite{city} has 19 annotated object categories in street scenes from 50 different cities. It contains 5,000 annotated images in which $2,975$ images for training, 500 images for validation, and 1,525 images for testing. The images are in high resolution with the size being $2048\times 1024$.
    \item \textbf{ADE20K} \cite{ade20k} is a challenging benchmark with 150 annotated object categories, which has 20210 training pairs, 2000 validation pairs, and 3352 testing pairs.
    \item \textbf{PASCAL Context} \cite{voc} has 60 annotated object categories (59 classes and 1 background). It has 4998 training pairs and 5105 validation pairs.
\end{itemize}

As mentioned above, SETR~\cite{vtseg} is the upper-bound model of our TransCL in image domain in this application. Thus, we apply the same training strategy as SETR. Specifically, at the beginning of training, we utilize the pre-trained parameters of ViT-L-16~\cite{vit} in the image classification task to initialize parameters in TB. Then we use the default setting of public codebase mmsegmentation \cite{mmseg2020} to train our TransCL. Concretely, the input image is randomly resized with the ratio between $0.5$ and 2 and randomly cropped with the size being 768, 512, and 480 for Cityscapes, ADE20K, and Pascal Context, respectively. A random horizontal flipping is also applied during training. We train the network on these three benchmarks with the batch size being 8, 16, and 8, respectively. We adopt a polynomial learning rate decay schedule \cite{pspnet} and employ SGD as the optimizer during training. The momentum and weight decay are set to 0.9 and 0, respectively. We set the initial learning rate as 0.001 on ADE20K and Pascal Context and 0.01 on Cityscapes. The CSM, TB, and TH in TransCL are jointly optimized by a pixel-wise cross-entropy loss.

\textbf{Comparison Results.} In this application, we focus on comparing our proposed TransCL with its upper bound model~\cite{vtseg} and some state-of-the-art methods~\cite{apcnet,gffnet,ccnet,pspnet,twins} in image domain. The quantitative comparison of semantic segmentation on ADE20k~\cite{ade20k}, Cityscapes~\cite{city}, and Pascal Context~\cite{voc} are presented in Tab.~\ref{seg_ade}, Tab.~\ref{seg_city}, and Tab.~\ref{seg_voc}, respectively. Clearly, our method can achieve comparable performance to SETR~\cite{vtseg} and outperform some state-of-the-art image-domain methods with significantly fewer measurements. The visualization comparison on three benchmarks is presented in Fig.~\ref{im_seg}, presenting the attractive semantic segmentation results of our methods. Specifically, in scenarios with few objects (\textit{e.g.}, the first row and second row of Fig.~\ref{im_seg}), our TransCL can achieve comparable performance to~\cite{vtseg} with much fewer measurements. As presented in the third row of Fig.~\ref{im_seg}, our TransCL maintains superior performance in the challenging ADE20K benchmark, which contains 150 annotated object categories. 

\textbf{Performance with Arbitrary CS Ratios.} In semantic segmentation, we also study the performance of our TransCL in the challenging setting with arbitrary CS ratios. The block size is set as $16$$\times$$16$. Thus, our TransCL can handle 256 different CS ratios ($\gamma$$\in$$ [\frac{1}{256},1]$) with a single model. We present the segmentation performance in Fig.~\ref{ab_seg}, presenting that the performance rapidly improves with the increase of CS ratio and then becomes stable. There is only a slight performance decrease compared with the model trained with fixed CS ratios. Note that in the challenging ADE20K~\cite{ade20k}, our proposed method can still maintain stable performance with arbitrary CS ratios. In the extreme case with the CS ratio being $\frac{1}{256}$, our proposed TransCL can still achieve satisfactory results on these three benchmarks. These results fully demonstrate that our proposed TransCL can perform complex high-level vision tasks in the measurement domain with state-of-the-art performance and robustly handle arbitrary CS ratios with a single model.    

\section{Ablation Study and Discussions}
In this section, we focus on studying the effectiveness of different components in our TransCL and presenting some superior properties and interesting findings of our TransCL.

\begin{figure}[h]
\centering
\small 
\includegraphics[width=.95\columnwidth,height=7cm]{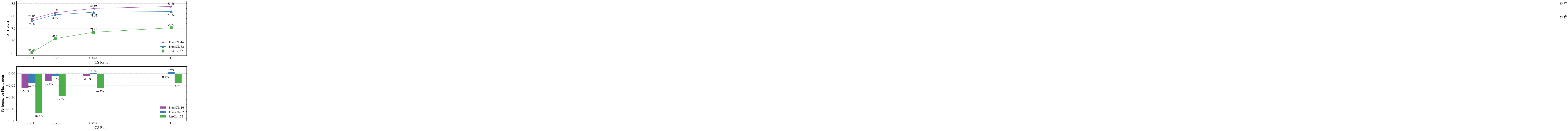}
\centering
\caption{Image classification performance comparison between our proposed TransCL and its CNN counterpart (ResCL) at different CS ratios and the performance changes compared with their upper bound model. The results reflect the advantages of our proposed TransCL in performance and stability.}
\label{ab1} 
\end{figure}

\subsection{Comparison with CNN Counterpart}
\label{5.1}
Non-local self-similarity is an effective prior in compensating for the information loss and resisting inference due to its robust capacity in constructing long-range correlations. To demonstrate its effectiveness in CL tasks, we compare our TransCL with its CNN counterpart. Specifically, we replace the transformer-based backbone (TB) in TransCL with the very deep ResNet-152~\cite{resnet} (represented as ResCL). We evaluate the performance of our TransCL and ResCL under various conditions in the image classification task. Note that the serialized input is intractable to the CNN architecture. Thus, we combine the input sequence into a square tensor in which each vectorized measurement is located in the position of the original image block.

\subsubsection{Performance and Stability}
The comparison is conducted on ImageNet 2012~\cite{imagenet} dataset with the CS ratios being $1\%$, $2.5\%$, $5\%$, and $10\%$. We report the top-1 ($\%$) classification accuracy of our TransCL and ResCL at different CS ratios and present the performance changes ($\%$) compared with their upper bound models (ViT-B~\cite{vit} and ResNet-152) in the first row and second row of Fig.~\ref{ab1}, respectively. Clearly, our TransCL enjoys advantages in both performance and stability. Specifically, our proposed TransCL performs better than ResCL in all evaluation points, and the margin becomes larger as the decrease of CS ratio. The stability of our TransCL is reflected in two aspects, \textit{i.e.}, the performance decays more slowly and is closer to the upper bound model.

\subsubsection{Resisting Noise Perturbation}
\begin{figure}[t]
\centering
\small 
\includegraphics[width=.95\columnwidth]{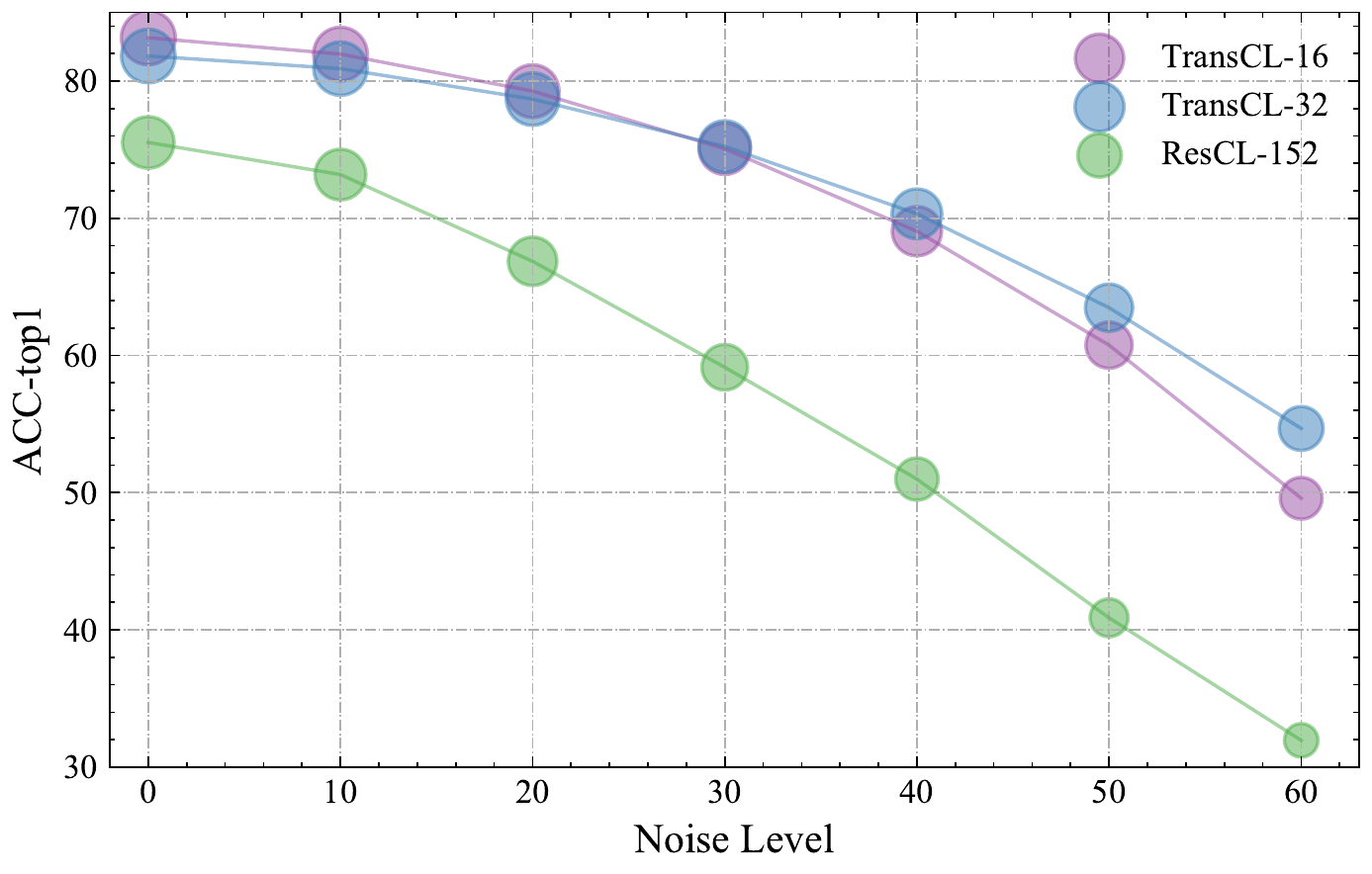}
\centering
\caption{The comparison of ability to resist noise perturbation of our TransCL and its CNN counterpart (ResCL). We present the top-1 ($\%$) accuracy on ImageNet validation set with the Gaussian noise level $\sigma \in \{0,10,20,...,60\}$.}
\label{noise} 
\end{figure}

In the process of CS, there usually exists noise perturbation from the natural environment and hardware. \cite{td2} exploited the performance bound of CL under the noise perturbation. In this part, we study the capacity of our TransCL and ResCL to resist the noise perturbation. Following \cite{td2}, we add random Gaussian noise $\mathbf{n}$$\sim$$ \mathcal{N}(\mu,\sigma^2)$ to original images. The CS process is correspondingly expressed as:
\begin{equation}
    \mathbf{y}_i=\mathbf{\Phi}_B^\gamma \mathbf{x}_i+\mathbf{\Phi}_B^\gamma \mathbf{n}.
\end{equation}
In the experiment, we utilize the zero-mean Gaussian noise and set the CS ratio as $10\%$. Note that all methods are trained with clean images, and we only add the noisy signal in the evaluation. We evaluate the performance under noise perturbation with different noise levels on the ImageNet validation set and present the comparison result in Fig.~\ref{noise}. Obviously, our proposed TransCL performs better than ResCL, demonstrating the effectiveness of the transformer-based backbone. We can also find that TransCL-32 performs better than TransCL-16, which can be interpreted by the fact that the TransCL-32 has a larger receptive field of matching patches leading to more robust long-range correlations to resist noisy signals.

\begin{figure}[t]
\centering
\footnotesize
\begin{minipage}[t]{0.3\linewidth}
\centering
\includegraphics[width=1\columnwidth,height=1\columnwidth]{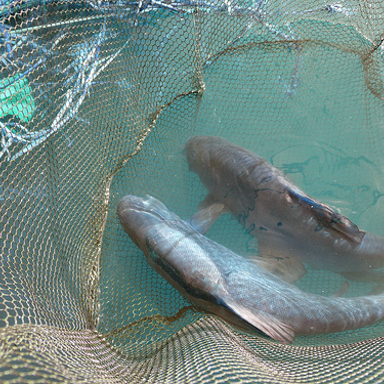}\\(a) Original Image
\end{minipage}
\begin{minipage}[t]{0.3\linewidth}
\centering
\includegraphics[width=1\columnwidth,height=1\columnwidth]{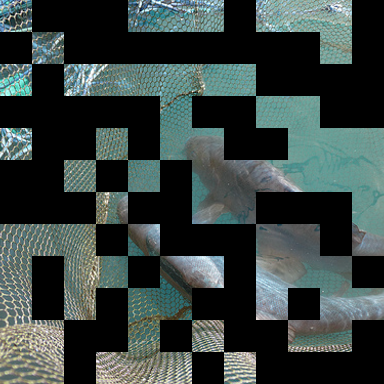}\\(b) Random PatchDrop
\end{minipage}
\begin{minipage}[t]{0.3\linewidth}
\centering
\includegraphics[width=1\columnwidth,height=1\columnwidth]{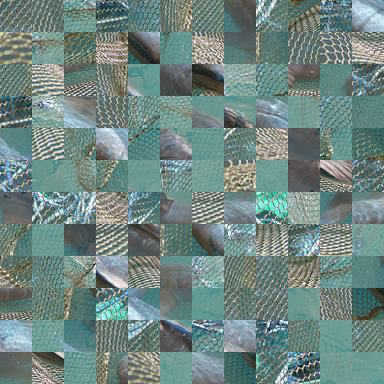}\\(c) Random PatchShuffle
\end{minipage}
\centering
\caption{Visualization of disturbing patterns, including random PatchDrop and random PatchShuffle.}
\label{im_ab3} 
\end{figure}

\begin{figure}[t]
\centering
\small 
\includegraphics[width=.95\columnwidth,height=5cm]{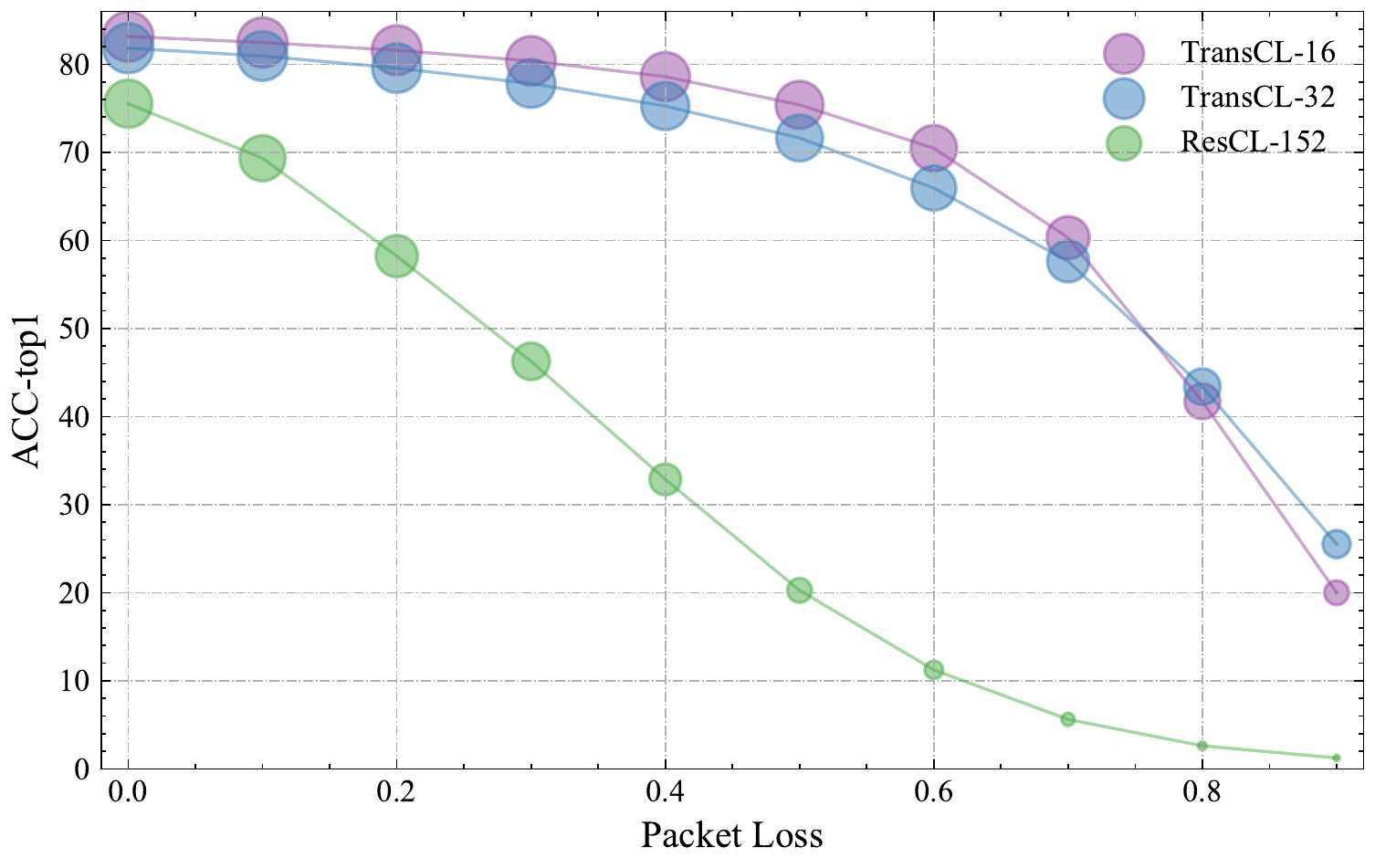}
\centering
\caption{The comparison of resisting random PatchDrop of our TransCL and its CNN counterpart (ResCL) on ImageNet validation set with various packet losses.}
\label{ab2} 
\end{figure}

\begin{table}[h]
    \centering
    \caption{Quantitative comparison of two-stage method and our single-stage TransCL on image classification task. We present the top-1 ($\%$) accuracy on ImageNet validation set in this table.}
    \footnotesize
    \begin{tabular}{c c c c c c}
    \bottomrule
    \multirow{2}{*}{Type}&\multirow{2}{*}{Method}&\multicolumn{4}{c}{CS ratio}\\
    \cline{3-6}
        & & $10\%$ & $5\%$ & $2.5\%$ & $1\%$  \\
        \bottomrule
         Two-stage &\tabincell{c}{OPINE-Net$^+$\cite{opin} \\ ViT-B-32\cite{vit}}& 80.65 & 77.00 & 70.89 & 55.57\\
         \hline
          Single-stage & \tabincell{c}{TransCL-32 \\ (Ours)} & \textbf{81.82} & \textbf{81.53} & \textbf{80.50} & \textbf{78.00}\\
         \hline
    \end{tabular}
    \label{ab_two}
\end{table}

\begin{figure}[h]
\centering
\small 
\includegraphics[width=.95\columnwidth]{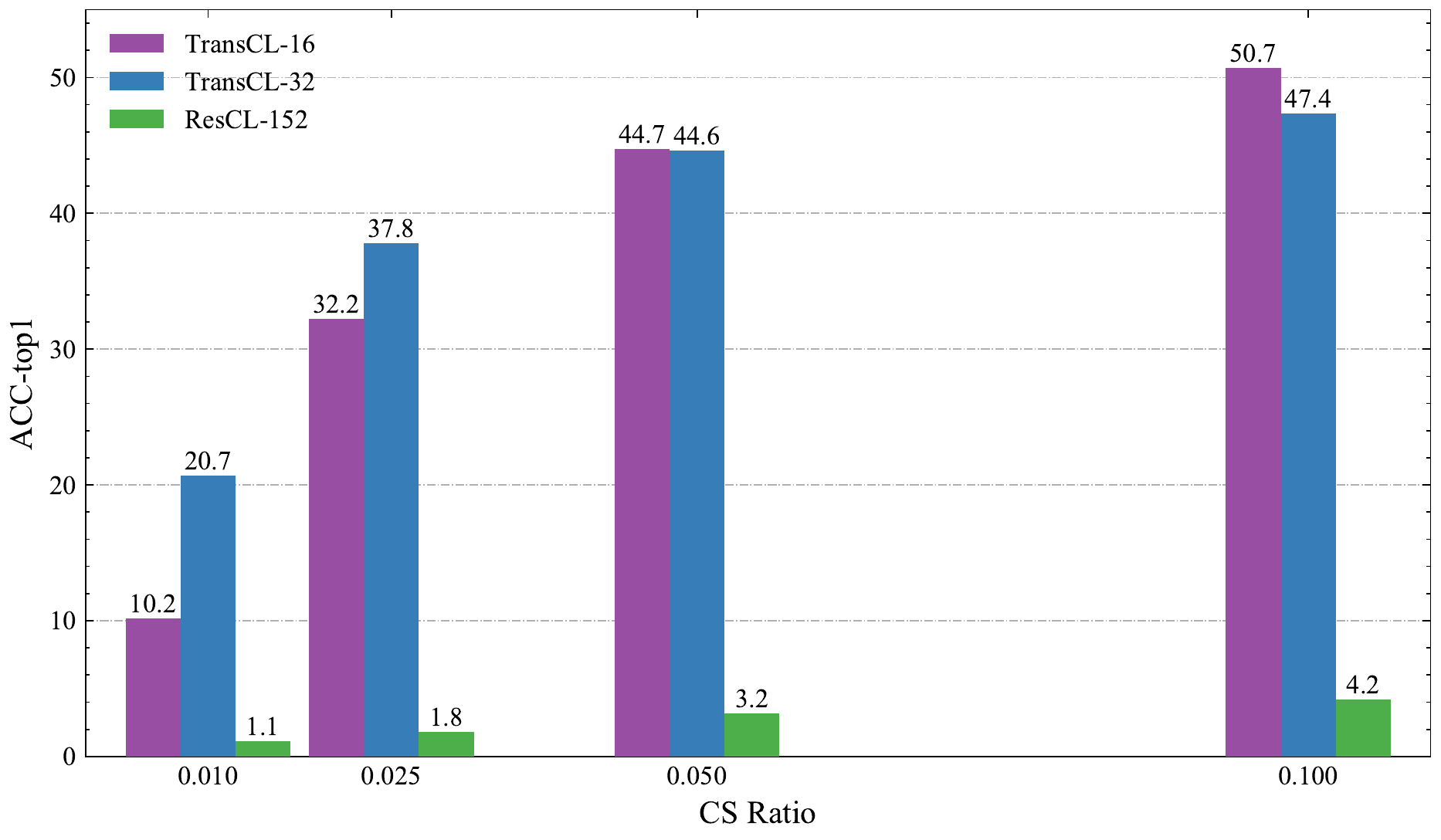}
\centering
\caption{Top-1 ($\%$) accuracy over ImageNet validation set with shuffled image patches. We compare our TransCL with its CNN counterpart (ResCL).}
\label{ab3} 
\end{figure}

\begin{table*}[h]
    \centering
    \caption{Comparison of complexity of various strategies to infer a $384\times 384$ image and the top-1 ($\%$) accuracy on ImageNet~\cite{imagenet} validation set. The additional complexity produced in measurement reconstruction stage is highlighted in \textcolor{red}{red}.}
    \footnotesize
    \begin{tabular}{c|c c c c c c c c}
         \hline
         Methods & ResCL-152 & ResCL-101 & ResCL-50 & DensCL-201 & DensCL-121 & Two Stages &  TransCL-16 & TransCL-32\\
         \hline
         \makecell[c]{Speed\\(ms/image)}  & 16.9 & 11.9 & 7.5 & 11.9 & 8.4 & 7.4$+$\textcolor{red}{25.2} & 30.15 & $\mathbf{7.4}$\\
         \hline
         \makecell[c]{Peak\\Memory (GB)} & 2.2 & 1.9 & 1.5 & 2.3 & 1.7 & $\max(1.5,\textcolor{red}{6.8})$ & 3.2 & \textbf{1.5}\\ 
         \hline
         Param. (M) & 60 & 45 & 25 & 20 & \textbf{8} & 88$+$\textcolor{red}{0.6} & 86 & 88\\
         \hline
         Flops (G) & 30.1 & 23.0 & 12.1 & 12.8 & \textbf{8.4} & 12.3$+$\textcolor{red}{112} & 49.3 & 12.3 \\
         \hline
         Accuracy ($\%$) & 75.22 & 71.77 & 70.19 & 72.05 & 69.23 & 80.65 & \textbf{83.36} & 81.82 \\
         \hline
    \end{tabular}
    \label{time}
\end{table*}
\subsubsection{Resisting Random PatchDrop and PatchShuffle}
To further demonstrate the superiority of our TransCL in resisting information loss and external interference, we deploy two types of interference, \textit{i.e.}, Random PatchDrop and Random PatchShuffle, which are only added during testing. As shown in Fig.~\ref{im_ab3}(b), for Random PatchDrop, we randomly mask some square regions of the block-based compressed sensing (BCS) process with a certain proportion to simulate the packet loss during data transmission. The evaluation results are presented in Fig.~\ref{ab2}, presenting that our proposed TransCL performs better than the ResCL with a large margin. This benefits from the fact that transformer can fully utilize the non-local property in sequence, rather than operating within a local region. For Random PatchShuffle, we eliminate the structural information within images by defining a shuffling operation on input image patches. One sample of this operation is presented in Fig.~\ref{im_ab3}(c), whose spatial structure is badly damaged. This setting can simulate data transmission in the wrong sequence. The evaluation results are presented in Fig.~\ref{ab3}, from which one can observe that the resistance ability of our TransCL to this interferential factor is obviously better than ResCL.  

\subsection{Comparison with Two-stage Method}
As illustrated in Fig.~\ref{cl}, a straightforward method to solve high-level vision tasks with CS measurements is reconstructing first and then inferring. However, this strategy potentially discloses private information and has computational redundancy, \textit{i.e.}, in many applications, we are not interested in obtaining a precise reconstruction of the scene under view, but rather are only interested in the results of inference tasks. In this part, we provide experimental results to demonstrate that the reconstruction step is almost unnecessary, and our proposed TransCL can achieve better performance with a single stage. In terms of the implementation of the two-stage method, we apply the top-performing method~\cite{opin} as the reconstruction module. Then we use ViT-B-32~\cite{vit} to perform image classification based on the reconstruction output. We set the CS ratio as $10\%$, $5\%$, $2.5\%$, and $1\%$. The comparison result on ImageNet validation set is presented in Tab.~\ref{ab_two}. One can see that the two-stage method has satisfied performance with high CS ratios, but the performance declines sharply with the decrease of CS ratio. It indicates that there exits serious coupling problem between the reconstruction model and inference model, \textit{i.e.}, the inference model trained with natural images can not adapt the input from CS restoration results. Therefore, for a satisfactory result, we need jointly optimize a specific restoration model and an inference model at each CS ratio, which is not feasible in practical applications. By contrast, our proposed TransCL can achieve attractive and stable performance without coupling problem while also having the advantage of privacy protection.     

\begin{table}[h]
    \centering
    \caption{Quantitative comparison of our compressive learning method with traditional data compression approaches on image classification task. We present the top-1 ($\%$) accuracy on ImageNet validation set.}
    \small
    \begin{tabular}{c c c c c}
    \bottomrule
         Sampling ratio & $10\%$ & $5\%$ & $2.5\%$ & $1\%$  \\
        \bottomrule
        \tabincell{c}{SVD+ViT-B-32\cite{vit} \\ (w/o re-training)}& 68.69 & 42.02 & 12.08 &  2.7\\
        \hline
         \tabincell{c}{Bicubic+ViT-B-32\cite{vit} \\ (w/o re-training)}& 74.00 & 68.24 & 60.12 & 42.51\\
         \hline
         Bicubic+ViT-B-32\cite{vit}& 81.00 & 80.03 & 78.48 & 73.91\\
         \hline
         TransCL-32 & \textbf{81.82} & \textbf{81.53} & \textbf{80.50} & \textbf{78.00}\\
         \hline
    \end{tabular}
    \label{bicubic}
\end{table}

\subsection{Complexity Analysis}
\label{s_comp}
As illustrated in Fig.~\ref{cl}, since no signal reconstruction is required, CL is an efficient way to perform high-level vision tasks in the measurement domain and should runs faster than two-stage methods (reconstruct and then infer). Additionally, the transformer-based backbone is efficient in data processing, which has been studied in ViT~\cite{vit} (please refer to Fig.11 in \cite{vit}). Thus, in this part, we focus on analyzing the model complexity of CL and two-stage strategy, and we compare our proposed TransCL with some CNN counterparts, \textit{i.e.}, ResCL and DensCL. Note that the implementation of two-stage method is the same as Tab.~\ref{ab_two}. In the experiment, we set the image size and CS ratio as $384$$\times$$384$ and $10\%$, respectively. The evaluation results of inference speed, peak memory, parameters, and Flops are presented in Tab.~\ref{time}, which are all evaluated on an NVIDIA TESLA T4 GPU. We also present the top-1 accuracy of each strategy on ImageNet validation set in this table. We can find that compared with some commonly used CNN-based backbones, our TransCL has advantages in terms of inference speed and Flops, and the efficiency is more obvious when compared with the two-stage strategy. Specifically, the time complexity and Flops of our TransCL-32 are significantly lower than the ResNet-152 and ResNet-101 and achieve comparable inference speed with the tiny ResNet-50. This property mainly benefits from the efficiency of dot products on GPU. In the two-stage strategy, the restoration model produces $3.4\times$ time complexity, $4.3\times$ peak memory, and $9.1\times$ flops even with much fewer parameters. This is because the reconstruction process is a pixel-to-pixel projection without changing spatial size. Thus, given a $384\times 384$ input, the reconstruction model demands a high computational burden. Due to the four times length of the input sequence, TransCL-16 has higher model complexity than TransCL-32. The performance gains are also obvious compared with other strategies.

We present the model complexity of different CL methods to measure a $384\times 384$ image with the CS ratio being $1\%$ in Tab.~\ref{complex}. We can find that VCL and MCL produce a lot of extra computational complexity in handling large-scale images. Such additional model complexity is unnecessary and makes them difficult to deal with real-world benchmarks.

\begin{table}[h]
    \centering
    \caption{Complexity comparison of various CL methods to sample a $384\times 384$ color image, with the CS ratio being $1\%$.}
    \vspace{4pt}
    \begin{tabular}{c c c c}
         \hline
          & VCL~\cite{vcl} & MCL~\cite{cl_w5} & TransCL (Ours)\\
         \hline
         Parameters & $2.0\times 10^9$ & $1.1\times 10^5$ & $\mathbf{2.7\times 10^3}$\\
         \hline
         Flops & $1.3\times 10^9$ & $4.6\times 10^7$ & $\mathbf{2.6\times 10^6}$\\
         \hline
    \end{tabular}
    \label{complex}
\end{table}

\subsection{Comparison with Traditional Data Compression Methods}
In this part, we want to show that learning in the measurement domain better preserves image information than the conventional data compression approaches. We compare our TransCL with the spatial downsampling and SVD~\cite{svd} methods. Specifically, for spatial downsampling, we utilize the bicubic algorithm to downsample the input image with the same ratio as the compressed sensing operation in our TransCL. Then, we utilize bilinear interpolation to resize the downsampled image to the original scale. To make a fair comparison, we also re-train a specific ViT-B model on ImageNet 2012 dataset for each downsampling factor, and the training strategy is the same as that we use to train our TransCL. For SVD compression, we decompose the input image by the SVD algorithm and then retain a specific amount of data according to the sampling ratio. Concretely, given an input image $\mathbf{I}\in \mathbb{R}^{w\times h}$, the SVD algorithm decomposes $\mathbf{I}$ as $\mathbf{I} = \mathbf{u}\mathbf{s} \mathbf{v}$,
where $\mathbf{u}\in \mathbb{R}^{w\times w}$ and $\mathbf{v}\in \mathbb{R}^{h\times h}$ are two orthogonal matrixes. $\mathbf{s} \in \mathbb{R}^{w\times h}$ is a diagonal matrix. For data compression, we can retain $k$ columns in $\mathbf{u}$ and $k$ rows in $\mathbf{v}$ according to the singular value in $\mathbf{s}$. The rebuilding result is defined as $\mathbf{I} = \mathbf{u}_{w\times k}\mathbf{s}_{k\times k} \mathbf{v}_{k\times h}$. The compression ratio is computed as $r=\frac{w\times k + h\times k+k}{w\times h}$. Because the SVD algorithm runs slowly on large-scale data, we only provide the performance of SVD without re-training. 

The comparison results are presented in Tab.~\ref{bicubic}. We can find that the special downsampling and SVD methods are not as efficient as compressive learning. Even though the joint optimization improves the performance of spatial downsampling, our method is still significantly better than the downsampling operation, and the superiority becomes more obvious with the decrease of sampling ratio. Furthermore, the spatial downsampling does not have the property of privacy protection. 

\begin{table}[t]
\caption{Comparison of the float-point sampling matrix and the binary sampling matrix. The result presents that our TransCL has the hardware-friendly property. The image classification task is conducted on TransCL-32, and the semantic segmentation task is conducted on TransCL-16.}
\centering
\footnotesize
\begin{tabular}{c c c c c c}
\bottomrule
 & CS Ratio & $1\%$ & $2.5\%$ & $5\%$ & $10\%$\\
\bottomrule
\multirow{2}*{\tabincell{c}{Classification \\ (Top-1 accuracy)}} & Float-point & 78.00 & 80.50 & 81.53 & 81.82\\
& Binary & 76.12 & 78.90 & 80.11 & 80.91\\
\cline{2-6}
& $\Delta$ & -1.88 & -1.6 & -1.42 & -0.09\\
\hline
\multirow{2}*{\tabincell{c}{Segmentation \\ (mIoU)}} & Float-point & 50.83 & 54.03 & 55.29 & 55.52\\
& Binary & 49.80 & 53.76 & 54.87 & 55.47\\
\cline{2-6}
& $\Delta$ & -1.03 & -0.27 & -0.42 & -0.05\\
\hline
\end{tabular}
\label{bin}
\end{table}

\begin{figure}[t]
\centering
\footnotesize
\begin{minipage}[t]{0.03\linewidth}
\centering
~\\
\makebox[\hh]{\rotatebox[origin=l]{90}{\makebox[\h][c]{\hspace{-\h}\normalsize{Learned float-point\ \ \ \ Fixed Gauss}}}}\\~\\~\\~\\~\\~\\~\\~\\~\\~\\~\\~\\~\\
\makebox[\hh]{\rotatebox[origin=l]{90}{\makebox[\h][c]{\hspace{-\h}\normalsize{Learned binary}}}}\\
\end{minipage}
\begin{minipage}[t]{0.3\linewidth}
\centering
\includegraphics[width=1\columnwidth,height=1\columnwidth]{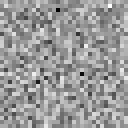}\vspace{0.2mm}\\
\includegraphics[width=1\columnwidth,height=1\columnwidth]{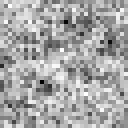}\vspace{0.2mm}\\
\includegraphics[width=1\columnwidth,height=1\columnwidth]{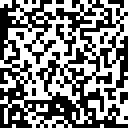}\\Spatial view
\end{minipage}
\begin{minipage}[t]{0.3\linewidth}
\centering
\includegraphics[width=1\columnwidth,height=1\columnwidth]{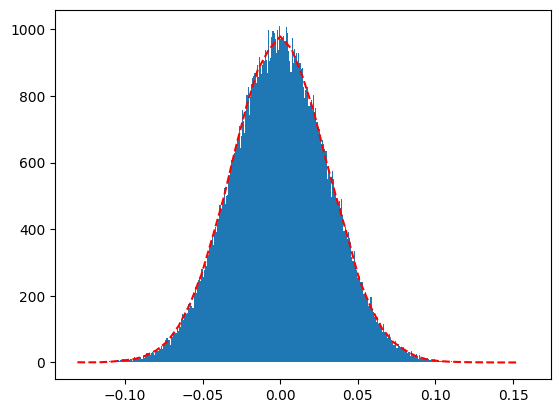}\vspace{0.2mm}\\
\includegraphics[width=1\columnwidth,height=1\columnwidth]{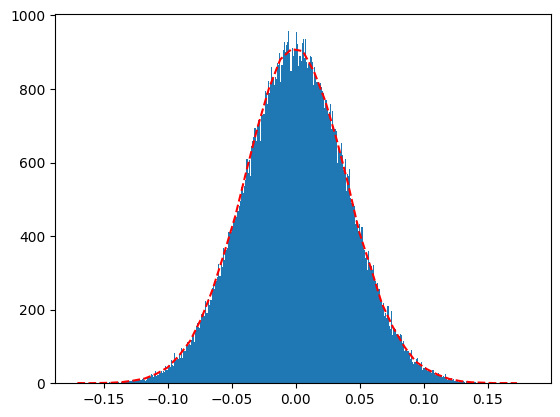}\vspace{0.2mm}\\
\includegraphics[width=1\columnwidth,height=1\columnwidth]{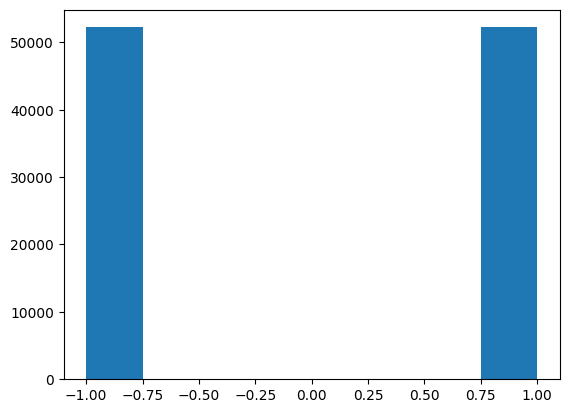}\\Distribution
\end{minipage}
\begin{minipage}[t]{0.3\linewidth}
\centering
\includegraphics[width=1\columnwidth,height=1\columnwidth]{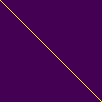}\vspace{0.2mm}\\
\includegraphics[width=1\columnwidth,height=1\columnwidth]{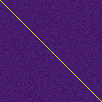}\vspace{0.2mm}\\
\includegraphics[width=1\columnwidth,height=1\columnwidth]{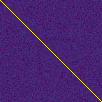}\\Orthogonality
\end{minipage}
\centering
\caption{Visualization of distribution and orthogonality of fixed Gaussian matrix (the first row), learned float-point matrix (the second row), and learned binary matrix (the third row).}
\label{ab_matrix} 
\end{figure}

\subsection{Sampling Matrix}
\subsubsection{Performance of Binary Sampling Matrix}
In this part, we make a performance comparison between learned float-point sampling and learned binary sampling matrix. The implementation is illustrated in Sec.~\ref{csm} and the experiments are conducted on image classification and semantic segmentation tasks with ImageNet 2012~\cite{imagenet} and Pascal Context~\cite{voc} datasets, respectively. As shown in Tab.~\ref{bin}, our proposed TransCL can achieve attractive performance with the binary sampling matrix, and there is only a slight performance reduction compared with using the float-point sampling matrix.

\subsubsection{Interesting Findings}
There are also some interesting findings of the sampling matrix. We take the CS ratio of 0.1 as an example. We reshape one row in a fixed random Gaussian matrix, our learned float-point matrix, and our learned binary matrix into $B$$\times$$B$ and visualize their spatial property in the first column of Fig.~\ref{ab_matrix}. Moreover, we visualize the distribution and orthogonality $\mathbf{\Phi}_B^\gamma (\mathbf{\Phi}_B^{\gamma})^\top$ of the sampling matrix in the second column and the third column of Fig.~\ref{ab_matrix}, respectively. One can see that the learned float-point sampling matrix obeys the Gaussian distribution, which is similar to the commonly used fixed random Gaussian matrix in many previous works \cite{cs_gaus1,cs_w1,cs_gaus3}. The learned binary matrix has an even distribution at the points of $-1$ and $1$. Fig.~\ref{ab_matrix} also presents that both float-point and binary sampling matrix is approximately orthogonal, \textit{i.e.}, $\mathbf{\Phi}_B^\gamma (\mathbf{\Phi}_B^\gamma)^\top$$\approx$$ \lambda \mathbf{I}$, where $\mathbf{I}$$\in$$ \mathbb{R}^{M_B^\gamma\times M_B^\gamma}$ is the identity matrix. This property is also consistent with the commonly used fixed random Gaussian matrix. Note that these properties are all learned without specific constraints. These similar characters naturally lead to the following inference: the learned sampling matrix retains the same properties as the fixed random Gaussian matrix, such as RIP~\cite{rip}. 

\begin{figure}[t]
\centering
\small
\begin{minipage}[t]{.8\linewidth}
\centering
\includegraphics[width=1\columnwidth,height=4.5cm]{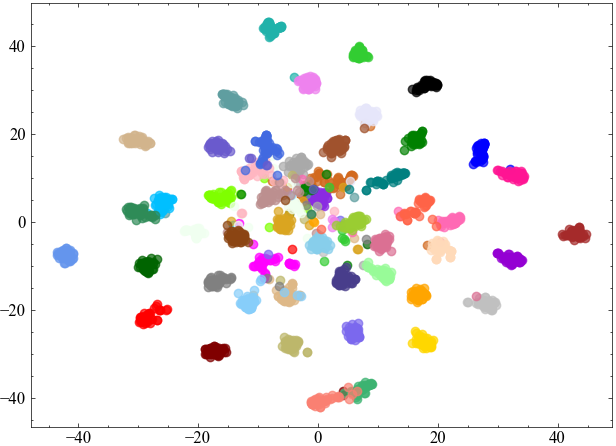}\\Feature representation in image domain (by ViT-B-32~\cite{vit})\\
\end{minipage}
\begin{minipage}[t]{.8\linewidth}
\centering
\includegraphics[width=1\columnwidth,height=4.5cm]{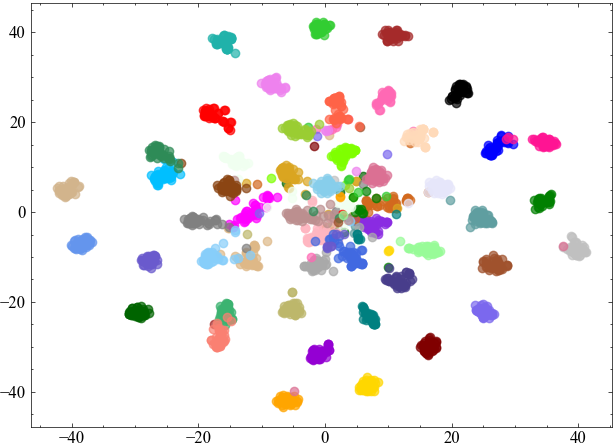}\\Feature representation in measurement domain (by our TransCL-32)\\
\end{minipage}
\centering
\caption{Visualization of feature representation in measurement domain and image domain. Each figure shows the feature representation after dimension reduction using the T-SNE method~\cite{tsne}.}
\label{feature_rep} 
\end{figure}

\subsection{Discussion: Feature Representation in Measurement Domain and Image Domain}
We generally assume that a large margin between measurement domain and image domain exists, and the performance of existing CL methods is not comparable to the image-domain methods. In this paper, we propose a novel CL method to bridge this gap. To further demonstrate our motivation, we visualize the feature representations from the measurement domain and the image domain, respectively. Concretely, we randomly select $50$ categories from $1000$ categories of the ImageNet validation set, each contains $50$ samples. Then we feed these sets of images to pre-trained TransCL-32-10 and ViT-B-32~\cite{vit}, and we collect the features extracted by these two models. Since the class information is stored in the class token, we utilize T-SNE~\cite{tsne} to reduce the dimensionality of the class token feature of each sample to $2$ dimensions for illustration purposes. The results from our TransCL and ViT-B are shown in Fig.~\ref{feature_rep}. We can find that the feature representations generated by our TransCL and ViT-B have good category discrimination, which demonstrates the superiority of our TransCL of better preserving image information from a few measurements. Fig.~\ref{feature_rep} also presents a high similarity of the feature representations from the measurement domain and image domain. Thus, these two domains are interrelated, and we can bridge the gap through specific designs.       

\section{Conclusion}
In this paper, we delve into the high-level visual applications of compressed sensing (CS) beyond the signal reconstruction, presenting that CS reconstruction is almost unnecessary in high-level vision tasks. Technically, we offer a completely new sequence-to-sequence perspective to compressive learning (CL) by applying the block-based compressed sensing (BCS) strategy, and we propose a novel transformer-based CL framework, dubbed TransCL. The proposed TransCL extends the concept of CL to more complex high-level vision tasks and larger datasets with real-world scale, which offers a fresh and successful instance and bridges the gap between inference tasks in the image domain and those in the measurement domain. Furthermore, we study several useful properties of our TransCL in terms of handling arbitrary CS ratios with a single model (Fig.~\ref{ab} and Fig.~\ref{ab_seg}), good generalization to binary sampling (Tab.~\ref{bin}), robustness under different degradation factors (Sec.~\ref{5.1}), privacy protection (Fig.~\ref{ab_pri}) and effective data compression (Tab.~\ref{bicubic}). Extensive experiments and theoretical analysis demonstrate that our proposed TransCL has these superior properties and can achieve state-of-the-art performance in image classification and semantic segmentation tasks, even at extremely low CS ratios. Our future work will focus on designing more efficient sequence-to-sequence CL methods and establishing real-world systems integrated with the hardware.

{
\bibliographystyle{IEEEtran}
\bibliography{reg}
}
\begin{IEEEbiography}[{\includegraphics[width=1in,height=1.25in,clip,keepaspectratio]{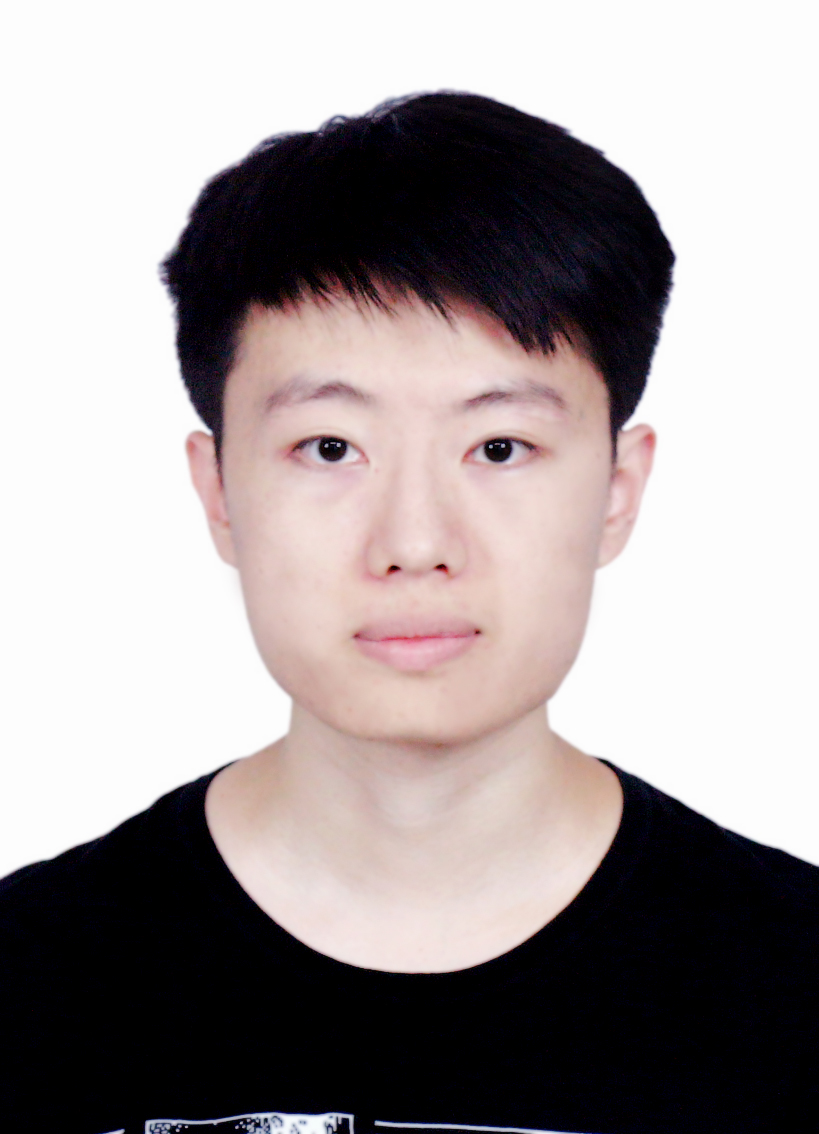}}]{Chong Mou} is currently pursuing the M.S. degree at the School of Electronic and Computer Engineering, Peking University Shenzhen Graduate School, Shenzhen, China. He received the B.S. degree at South China University of Technology, in 2020. His research interests include image denoising, feature attention, pattern recognition and deep learning.
\end{IEEEbiography}

\begin{IEEEbiography}[{\includegraphics[width=1in,height=1.25in,clip,keepaspectratio]{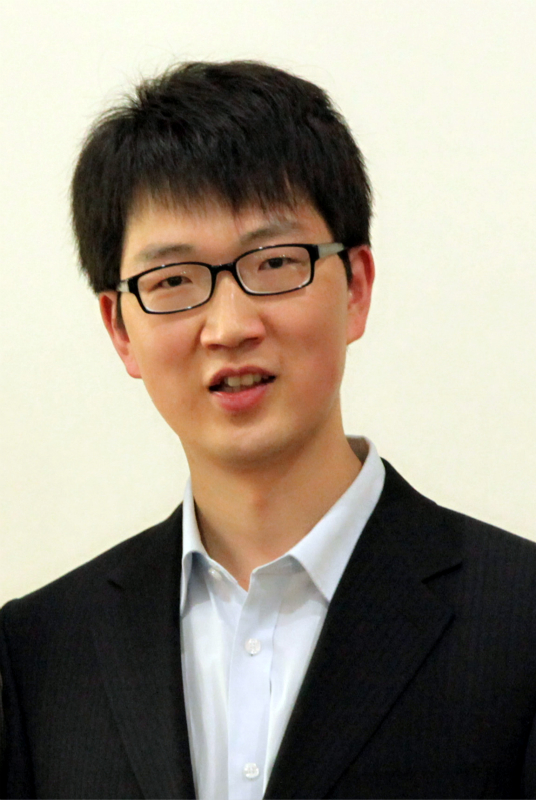}}]{Jian Zhang} (M'14) received the B.S. degree from the Department of Mathematics, Harbin Institute of Technology (HIT), Harbin, China, in 2007, and received his M.Eng. and Ph.D. degrees from the School of Computer Science and Technology, HIT, in 2009 and 2014, respectively. From 2014 to 2018, he worked as a postdoctoral researcher at Peking University (PKU), Hong Kong University of Science and Technology (HKUST), and King Abdullah University of Science and Technology (KAUST). 

Currently, he is an Assistant Professor with the School of Electronic and Computer Engineering, Peking University Shenzhen Graduate School, Shenzhen, China. His research interests include intelligent multimedia processing, deep learning and optimization. He has published over 90 technical articles in refereed international journals and proceedings. He received the Best Paper Award at the 2011 IEEE Visual Communications and Image Processing (VCIP) and was a co-recipient of the Best Paper Award of 2018 IEEE MultiMedia.
\end{IEEEbiography}

\end{document}